\documentclass[runningheads]{llncs}

\usepackage{eccv}

\usepackage{eccvabbrv}

\usepackage{graphicx}
\usepackage{booktabs}
\usepackage{algorithm, algpseudocode, setspace, multirow}

\usepackage[accsupp]{axessibility}  

\newcommand{\Ac}{{\mathcal A}}

\newcommand{\Dc}{{\mathcal D}}
\newcommand{\Ec}{{\mathcal E}}

\newcommand{\Nc}{{\mathcal N}}

\newcommand{\Rc}{{\mathcal R}}

\newcommand{\Lc}{{\mathcal L}}
\newcommand{\Mc}{{\mathcal M}}
\newcommand{\veps}{\boldsymbol{\epsilon}}
\newcommand{\rmI}{\boldsymbol{I}}
\def\eqref#1{(\ref{#1})}

\newcommand{\x}{{\boldsymbol x}}

\newcommand{\y}{{\boldsymbol y}}
\newcommand{\z}{{\boldsymbol z}}

\newcommand{\Ed}{{\mathbb E}}

\newcommand{\etab}{{\boldsymbol \eta}}

\DeclareMathOperator*{\argmin}{arg\,min}

\usepackage[pagebackref,breaklinks,colorlinks,citecolor=eccvblue]{hyperref}

\usepackage{orcidlink}

\begin{document}

\title{DreamSampler: Unifying Diffusion Sampling and Score Distillation for Image Manipulation} 

\titlerunning{DreamSampler}

\author{Jeongsol Kim\inst{1, *}\orcidlink{0000-0002-1700-4548} \and
Geon Yeong Park\inst{1, *}\orcidlink{0009-0006-7522-4553} \and
Jong Chul Ye\inst{2}\orcidlink{0000-0001-9763-9609}}

\authorrunning{Kim \& Park et al.}


\institute{Dept. of Bio \& Brain Engineering, KAIST\and
Kim Jae Chul AI graduate school, KAIST\\
\email{\{jeongsol, pky3436, jong.ye\}@kaist.ac.kr}\\
* Equal contribution}
\maketitle

\begin{figure}[h]
    \centering
    \includegraphics[width=\linewidth]{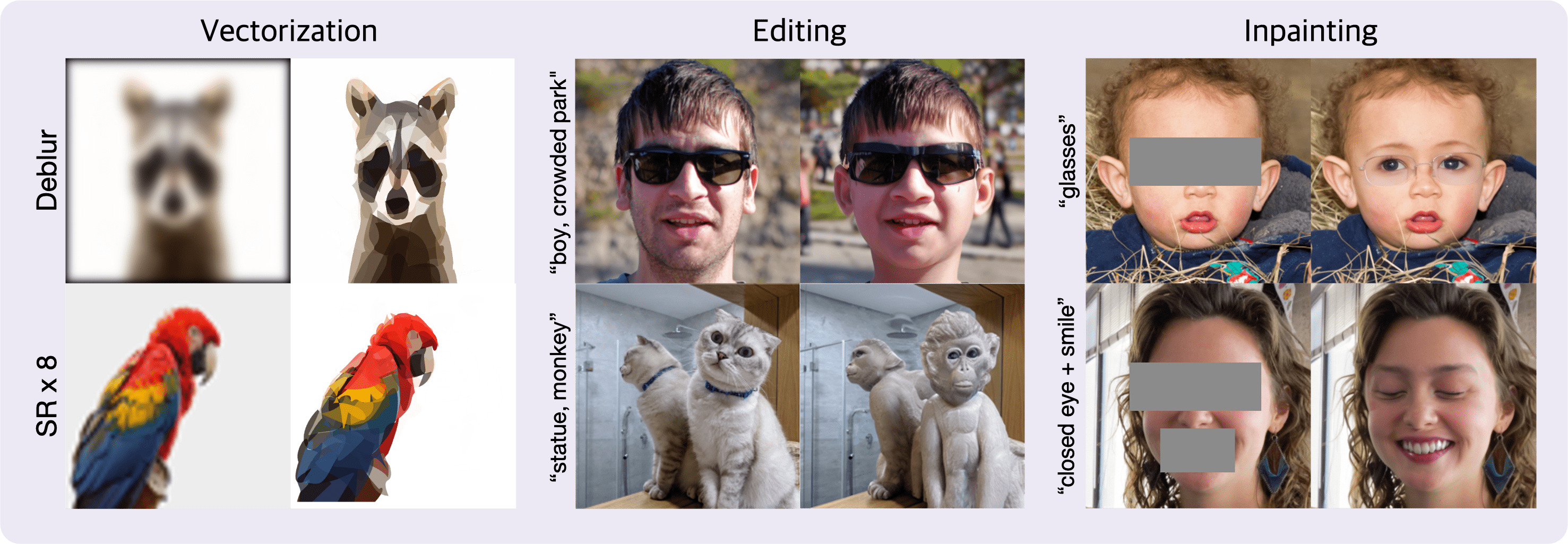}
    \caption{DreamSampler can be used for vectorized image restoration, editing, text-guided inpainting, etc. Code: \url{https://github.com/DreamSampler/dream-sampler}}
\end{figure}
\vspace{-1cm}

\begin{abstract}
Reverse sampling and score-distillation have emerged as main workhorses in recent years for image manipulation using latent diffusion models (LDMs). 
While reverse diffusion sampling often requires adjustments of LDM architecture or feature engineering, score distillation offers a simple yet powerful model-agnostic approach, but it is often prone to mode-collapsing.
To address these limitations and leverage the strengths of both approaches, here we introduce a novel framework called {\em DreamSampler}, which seamlessly integrates these two distinct approaches through the lens of regularized latent optimization. Similar to score-distillation, DreamSampler is a model-agnostic approach applicable to any LDM architecture, but it allows both distillation and reverse sampling with additional guidance for image editing and reconstruction.
Through experiments involving image editing, SVG reconstruction and etc, we demonstrate the competitive performance of DreamSampler compared to existing approaches, while providing new applications. 
\keywords{Latent diffusion model \and Generation \and Score distillation}
\end{abstract}

\section{Introduction}
\label{sec:intro}

\begin{figure}[h]
    \centering
    \includegraphics[width=\linewidth]{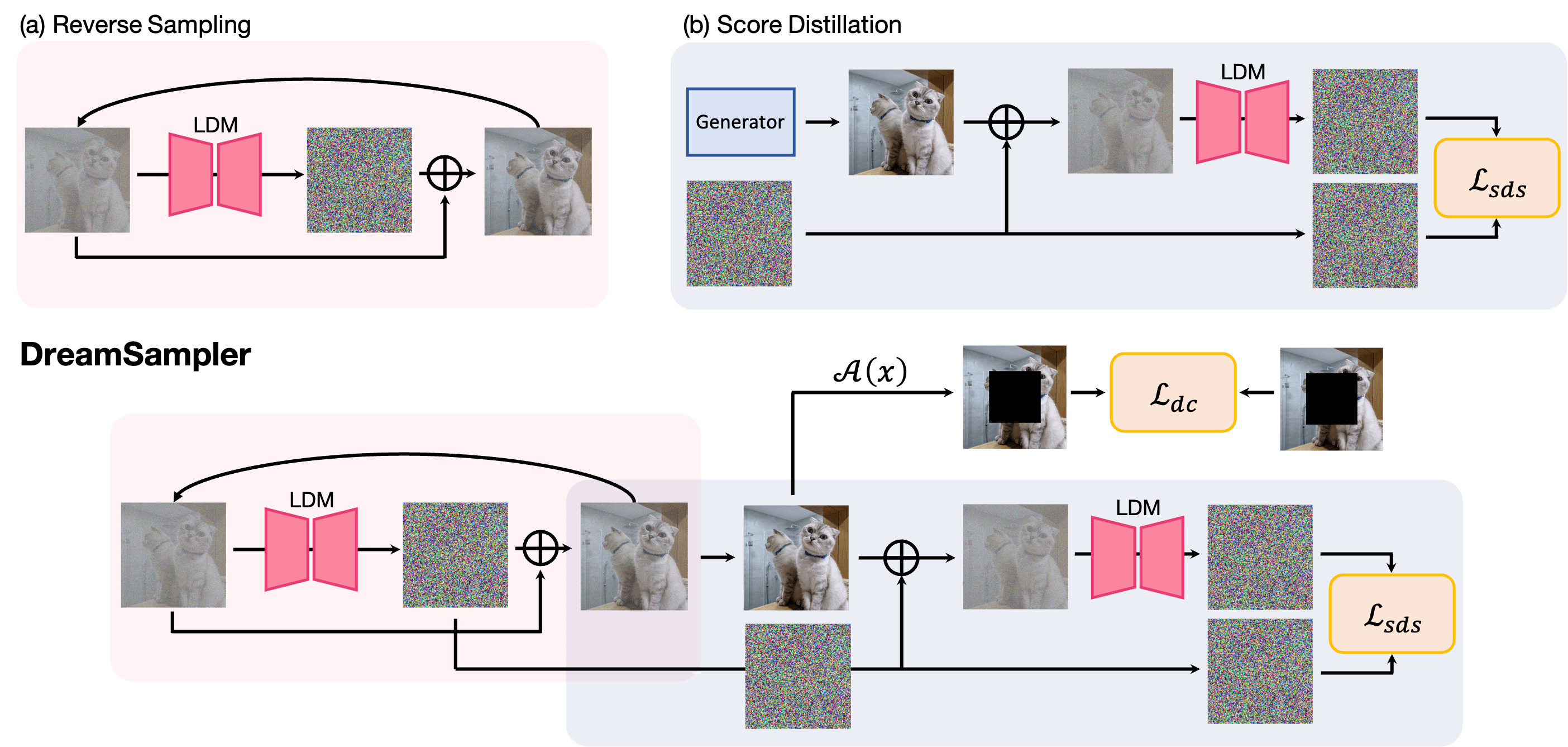}
    \caption{DreamSampler vs (a) reverse diffusion and (b) score distillation.}
    \label{fig:overview}
\end{figure}

Diffusion models~\cite{ho2020denoising, song2020score, song2020denoising} have been extensively studied as powerful generative models in recent years. 
These models operate by generating clean images from Gaussian noise through a process termed {ancestral sampling}. This involves progressively reducing noise by utilizing an estimated score function to guide the generation process from a random starting point towards the distribution of natural images.
Reverse diffusion sampling introduces stochasticity through the reverse Wiener process within the framework of SDE~\cite{song2020score}, contributing to the prevention of mode collapse in generated samples while enhancing the fidelity~\cite{dhariwal2021diffusion}.
Furthermore, the reverse diffusion can be flexibly regularized by various guidance gradients. For instance, classifier-guidance~\cite{dhariwal2021diffusion} applies classifier gradients to intermediate noisy samples, facilitating conditional image generation. Additionally, DPS~\cite{chung2022diffusion} utilizes approximated likelihood gradients to constrain the sampling process, ensuring data consistency between the current estimated solution and given observation, thereby solving noisy inverse problems in a zero-shot manner.

On the other hand, another type of approach, called  \textit{score distillation}, 
utilizes diffusion model as prior knowledge for image generation and editing. For example, DreamFusion~\cite{poole2022dreamfusion} leveraged 2D text-conditioned diffusion models for text-guided 3D representation learning via NeRF.  
Here, the diffusion model serves as the teacher model, generating gradients by comparing its predictions with the label noises and guiding the generator as the student model.
The strength of the score distillation method lies in its ability to leverage the pre-trained diffusion model in a black-box manner without requiring any feature engineering, such as adjustments to the model architecture.
Unfortunately, the score distillation is more often prone to mode collapsing compared to the reverse diffusion.

Although both algorithms are grounded in the same principle of diffusion models, the approaches appear to be different, making it unclearly how to synergistically combine the two approaches.
To address this, we introduce a unified framework called {\em DreamSampler}, that seamlessly integrates two distinct approaches and take advantage of the both worlds through the lens of regularized latent optimization.
Specifically, DreamSampler is model-agnostic and does not require any feature engineering, such as adjustments to the model architecture. Moreover, in contrast to the score distillation, DreamSampler is free of mode collapsing thanks to the stochastic nature of the sampling.

The pioneering aspect of DreamSampler is rooted in two pivotal insights. 
First, we demonstrate that the process of latent optimization during reverse diffusion can be viewed as a proximal update from the posterior mean by Tweedie's formula. This interpretation allows us to integrate additional regularization terms, such as measurement consistency in inverse problems, to steer the sampling procedure. Moreover, we illustrate that the loss associated with the proximal update can be conceptualized as the score distillation loss. 
This insight bridges a natural connection between the score-distillation methodology and reverse sampling strategies, culminating in their harmonious unification. In subsequent sections, we will explore various applications emerging from this integrated framework and demonstrate the efficacy of DreamSampler through empirical evidence.

\section{Motivations}
\label{sec:method}

\subsection{Preliminaries}
In LDMs\cite{rombach2022high}, the encoder $\Ec_\phi$ and the decoder $\Dc_\varphi$ are trained as auto-encoder,  satisfying $\x=\Dc_\varphi(\Ec_\phi(\x))=\Dc_\varphi(\z_0)$ where $\x$ denotes clean image and $\z_0$ denotes encoded latent vector. Then, the diffusion process is defined on the latent space, which is a range space of $\Ec_\phi$. Specifically, the forward diffusion process is 
\begin{align}
    \z_t = \sqrt{\bar\alpha_t} \z_0 + \sqrt{1-\bar\alpha_t} \veps
\end{align}
\noindent
where $\bar\alpha_t$ denotes pre-defined coefficient that manages noise scheduling, and $\veps \sim \Nc(0, \rmI)$ denotes a noise sampled from normal distribution.
The reverse diffusion process requires a score function via a neural network (i.e. diffusion model, $\veps_\theta$) trained by denoising score matching~\cite{ho2020denoising, song2020score}:
\begin{align}
    \min_\theta \Ed_{t, \veps\sim\Nc(0, \rmI)} \|\veps - \veps_\theta(\z_t, t)\|^2_2.
\end{align}
According to formulation of DDIM~\cite{song2020denoising, chung2023fast}, the reverse sampling from the posterior distribution $p(\z_{t-1}|\z_t, \z_0)$ could be described as
\begin{align}
    \z_{t-1} &= \sqrt{\bar\alpha_{t-1}} \hat\z_{0|t} + \sqrt{1-\bar\alpha_{t-1}}\tilde\veps \label{eqn:add_noise}
\end{align}
\noindent where
\begin{align}
    \hat\z_{0|t} &= (\z_t - \sqrt{1-\bar\alpha_t} \veps_\theta(\z_t, t)) / \sqrt{\bar\alpha_t} 
    \label{eqn:tweedie}\\
    \tilde\veps &= \frac{\sqrt{1-\bar\alpha_{t-1}-\eta^2\beta_t^2}\veps_\theta(\z_t, t)}{\sqrt{1-\bar\alpha_{t-1}}} + \frac{\eta \beta_t \veps}{\sqrt{1-\bar\alpha_{t-1}}}.
    \label{eqn:noise}
\end{align}
Here, $\hat\z_{0|t}$ refers to the denoised latent through Tweedie's formula,
and $\tilde\veps$ is the noise term composed of both deterministic $\veps_\theta(\z_t, t)$ and stochastic term
$\veps \sim \Nc(0, \rmI)$. 
Note that $\eta$ and $\beta_t$ denote variables that controls the stochastic property of sampling. When $\eta\beta_t=0$, the sampling is deterministic.

For text conditioning, classifier-free-guidance (CFG)~\cite{ho2022classifier} is widely leveraged. The estimated noise is computed by
\begin{align}  \label{eqn:cfg}
    \veps_\theta^\omega(\z_t, t, c_{ref}) = \veps_\theta(\z_t, t, c_\varnothing) + \omega [\veps_\theta(\z_t, t, c_{ref}) - \veps_\theta(\z_t, t, c_\varnothing)]
\end{align}
where $\omega$ denotes the guidance scale, $c_\varnothing$ refers to the null-text embedding, and $c_{ref}$ is the conditioning text embedding, which are encoded by pre-trained text encoder such as CLIP~\cite{radford2021learning}.
For simplicity, we will interchangeably use the terms $\veps_\theta(\z_t)$, $\veps_\theta(\z_t, t)$ and $\veps_\theta^\omega(\z_t, t, c_{ref})$ unless stated otherwise.

\subsection{Key Observations}
\label{sec:observations}
Score distillation sampling (SDS)~\cite{poole2022dreamfusion} and reverse diffusion~\cite{ho2020denoising, song2020denoising, song2020score} represent two distinct methodologies, each with its own pros and cons. SDS is an optimization method that focuses on minimizing score distillation loss, while reverse diffusion utilizes ancestral sampling, which stems from SDE or DDPM formulations. Although SDS is straightforward and model-agnostic, it often suffers from mode collapse due to its non-stochastic nature. Conversely, ancestral sampling typically avoids mode collapsing and generates more diverse outputs, but it is non-trivial to generalize the ancestral sampling with generic parameter space, e.g. NeRF MLP. The primary contribution of our study is the integration of these two approaches through an optimization perspective based on following key observations.
The first key insight of DreamSampler is that the
DDIM sampling can be interpreted as the solution of following optimization problem:
\begin{align}
    \z_{t-1} = \sqrt{\bar\alpha_{t-1}}\bar\z + \sqrt{1-\bar\alpha_{t-1}}\tilde\veps,\quad
    \mbox{where}\quad \bar\z = \argmin_\z \| \z - \hat\z_{0|t} \|^2 \label{eqn:obj_ddim}
\end{align}
Although looks trivial, one of the important implications of \eqref{eqn:obj_ddim} is that we can now extend the solution of \eqref{eqn:obj_ddim} to include additional regularization term,
\begin{equation}
    \min_\z \| \z - \hat\z_{0|t} \|^2_2 + \lambda_{reg}\Rc(\z)
    \label{eqn:obj_ddim_reg}
\end{equation}
where $\lambda_{reg}$ is scalar weight for the regularization function $\Rc(\z)$.
For example, we can use data consistency loss to ensure that the updated variable agrees with the given observation during inverse problem solving~\cite{kim2023regularization}.

Second key insight arises from the important connection between  \eqref{eqn:obj_ddim} and the score-distillation loss.
Specifically, using the forward diffusion to generate $\z_t$ from the clean latent $\z$:
\begin{align}
    \z_t &= \sqrt{\bar\alpha_t}\z + \sqrt{1-\bar\alpha_t}\veps, 
\end{align}
the objective function in \eqref{eqn:obj_ddim} can be converted to:
\begin{align}
    \| \z - \hat\z_{0|t} \|^2 &= \left\| \frac{\z_t-\sqrt{1-\bar\alpha_t}\veps}{\sqrt{\bar\alpha_t}} - \frac{\z_t-\sqrt{1-\bar\alpha_t}\veps_\theta(\z_t, t)}{\sqrt{\bar\alpha_t}} \right\|^2\\
    &= \frac{1-\bar\alpha_t}{\bar\alpha_t} \|\veps -\veps_\theta(\z_t, t) \|^2,
\end{align}
which is equivalent to the score-distillation loss up to a constant scaling factor.

\begin{figure}[t]
    \centering
    \includegraphics[width=\linewidth]{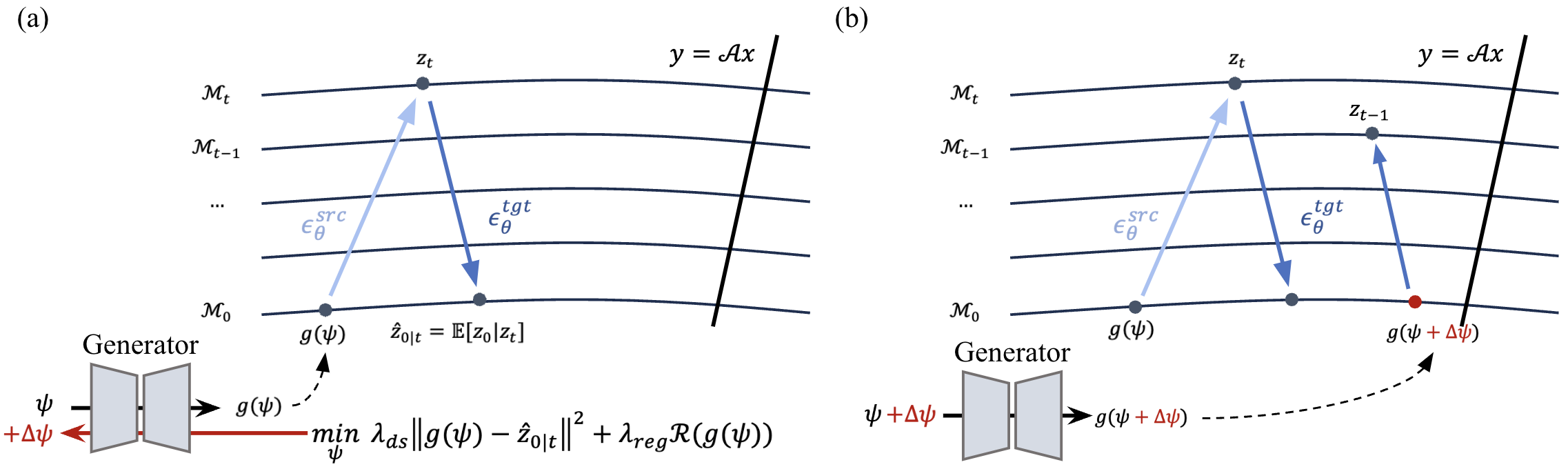}
    \caption{Unified framework of DreamSampler. (a) Distillation step where the gradient is computed from regularized latent optimization problem. (b) Reverse sampling step where estimated noise by diffusion model is added to the updated generation.}
    \label{fig:geometry}
\end{figure}

\vspace{-0.4cm}
\section{DreamSampler}
From Section \ref{sec:observations}, it is evident that DDIM sampling inherently includes `score-distillation' optimization, although it is conducted with $\z$ instead of generic parameters $\psi$. Inspired by this observation, we aim to generalize this optimization-based sampling with arbitrary generic parameter $\psi$, as in conventional score distillation sampling protocols.

\vspace{-0.3cm}
\subsection{General Formulation}
Suppose that $g(\psi)$ denotes a generated data by an arbitrary generator $g$ and parameter $\psi$.
Inspired by the two key insights described in the previous section, the sampling process of DreamSampler at timestep $t$ is given by\footnote{Here, we omit the encoder $\Ec_\phi$ in $\Ec_\phi(g(\psi))$ for notational simplicity. The encoder maps the generated image $g(\psi)$ to the latent space.}
\begin{align}
    \z_{t} &= \sqrt{\bar\alpha_{t}} g(\psi_{t}) + \sqrt{1-\bar\alpha_{t}} \tilde\veps,
\end{align}
where the noise $\tilde\veps$ is defined as in \eqref{eqn:noise},
and
\begin{align}
    \psi_t &= \argmin_\psi \| g(\psi)-\hat\z_{0|t} \|^2 + \lambda_{reg}\Rc(g(\psi)), \\
    \hat\z_{0|t} &=\left(\z_t - \sqrt{1-\bar\alpha_t}\veps_\theta(\z_t, t)\right) / \sqrt{\bar\alpha_t}.
    \label{eqn:our_optim}
 \end{align}
It implies that the score distillation sampling and reverse sampling can be integrated based on this generalized latent optimization framework with proper generator and regularization functions. In the following sections, we further delineate the special cases of DreamSampler.

\subsection{DreamSampler with External Generators}
Similar to DreamFusion \cite{poole2022dreamfusion}, for any differentiable generator $g$, DreamSampler can feasibly update parameters $\psi$ by leveraging the diffusion model and sharing the same sampling process.
To emphasize the distinctions between the original score distillation algorithms and DreamSampler, we conduct a line-by-line comparison of the pseudocode in Algorithm~\ref{alg:distillation} and Algorithm~\ref{alg:caseA}.

\begin{figure}[t]
\begin{minipage}{.49\textwidth}
    \vspace{-0.7cm}
    \begin{algorithm}[H]
    \setstretch{1.06}
            \small
           \caption{Score Distillation}
           \label{alg:distillation}
            \begin{algorithmic}[1]
             \Require $T$, $\zeta$, $g$, $\psi$, $\Ec_\phi$, {\{$\bar\alpha_t\}_{t=1}^T$}
             \State $\z_0 \gets \Ec_\phi(\x_0)$
              \For{$i=T$ {\bfseries to} $1$}
                 \State{\color{purple}$t \sim U[0, T]$}
                 \State{\color{teal}$\tilde\veps \sim \Nc(0, \rmI)$}
                 \State{$\z_t \gets \sqrt{\bar\alpha_t}\z_0 + \sqrt{1-\bar\alpha_t}\tilde\veps$}
                 \State{$\hat\veps_\theta \gets \veps_\theta^\omega(\z_t, t, c)$}
                 \State{$\nabla_\psi \Lc_{ds} \gets \tilde\veps - \hat\veps_\theta$}
                 \State{$\psi \gets \psi - \zeta\nabla_\psi \Lc_{ds}$}
                 \State{$\z_0 \gets \Ec_\phi(g(\psi))$}
              \EndFor
              \State {\bfseries return} $\psi$
            \end{algorithmic}
    \end{algorithm}
\end{minipage}
\begin{minipage}{.49\textwidth}
    \vspace{-0.7cm}
    \begin{algorithm}[H]
            \small
           \caption{DreamSampler}
           \label{alg:caseA}
            \begin{algorithmic}[1]
             \Require $T$, $\zeta$, $g$, $\psi$, $\Ec_\phi$, {\{$\bar\alpha_t\}_{t=1}^T$}
              \State $\z_0 \gets \Ec_\phi(\x_0), \veps_\theta(\z_{T+1}):= \veps\sim \Nc(0, \rmI)$
              \For{$i=T$ {\bfseries to} $1$}
                 \State{\color{purple}$t \gets i$, $\veps\sim\Nc(0,\rmI)$}
                 \State{\color{teal}$\tilde\veps \gets \frac{\sqrt{1-\bar\alpha_{t-1}-\eta^2\beta_t^2}\hat\veps_\theta + \eta\beta_t \veps}{\sqrt{1-\bar\alpha_t}}$}
                 \State{$\z_t \gets \sqrt{\bar\alpha_t}\z_0 + \sqrt{1-\bar\alpha_t}\tilde\veps$}
                 \State{$\hat\veps_\theta \gets \veps_\theta^\omega(\z_t, t, c)$}
                 \State{$\nabla_\psi \Lc_{ds} \gets \tilde\veps - \hat\veps_\theta$}
                 \State{$\psi \gets \psi - \zeta [\nabla_\psi \Lc_{ds} + \textcolor{cyan}{\lambda_{reg} \nabla_\psi \Rc(\z)} ]$}
                 \State{$\z_0 \gets \Ec_\phi(g(\psi))$}
              \EndFor
              \State {\bfseries return} $\psi$
            \end{algorithmic}
    \end{algorithm}
\end{minipage}
\vspace{-0.5em}
\end{figure}

First, DreamSampler follows the timestep schedule of the reverse sampling process, while distillation sampling algorithms use uniformly random timestep for optimization. This provides us with a novel potential for further refinement in utilizing various time schedulers to improve reconstruction quality or accelerate sampling. Second, as DreamSampler is built upon the general proximal optimization framework, it is compatible with additional regularization functions. From this design, one can explore various applications of the proposed distillation sampling. For example, by defining the regularization function as a data consistency term, one can constrain the generator to reconstruct the true image that aligns with the given measurement for inverse imaging.
 
Figure~\ref{fig:geometry} illustrates the sampling process of DreamSampler. 
At each timestep, the generated image $g(\psi)$ is mapped to a noisy manifold by incorporating the estimated noise from the previous timestep, and new noise is subsequently estimated by the diffusion model. The distillation gradient is then computed between these two estimated noises and utilized to update the generator parameters. 

\subsection{DreamSampler for Image Editing}
\label{sec:method_caseB}

As DreamSampler is a general framework, we can  reproduce other existing algorithms  by properly defining $g(\psi)$, $\hat\z_{0|t}$, and $\Rc$. As a representative example, here we derive Delta Denoising Score (DDS)~\cite{hertz2023delta} for image editing task and
discuss its potential extension from the perspective of DreamSampler.

The main assumption of DDS is to decompose the SDS gradient~\cite{poole2022dreamfusion} into text component and bias component, where only the text component contains information to be edited according to given text prompt while the bias component includes preserved information.
To remove the bias component from distillation gradient, DDS leverages the difference of two conditional predicted noises, $\veps_\theta(\z_t, t, c_{tgt}) - \veps_\theta(\z_t, t, c_{src})$, where $c_{tgt}$ denotes description of editing direction and $c_{src}$ denotes description of the original image.  Specifically, DDS update to clean latent reads
\footnote{
The update term is equivalent to (3) in \cite{hertz2023delta} when $\theta=\z$, $\hat\y=c_{src}$, and $\y=c_{tgt}$.}
\begin{align}
    \bar\z = \z - \gamma [\veps_\theta(\z_t, t, c_{tgt}) - \veps_\theta(\z_t, t, c_{src})].
    \label{eqn:onestep_dds}
\end{align}
In the context of Dreamsampler, let the generator $g(\psi):=\z$ be a clean latent.
Then, the following Theorem~\ref{theorem:dds} shows that the one-step DDS update can be reproduced with DreamSampler, by defining the regularization function $\Rc(\z)$ as Euclidean distance from the posterior mean conditioned on text.

\begin{theorem}
\label{theorem:dds}
Supposed $c_{src}$ in \eqref{eqn:onestep_dds} be defined as the null-text, i.e. $c_{src}=c_\varnothing$
and consider text-conditioned posterior mean:
        \begin{align}
    \hat\z_{0|t}(c) = \Ed[\z_0|\z_t, c] = (\z_t - \sqrt{1-\bar\alpha_t}\veps_\theta(\z_t, t, c))/\sqrt{\bar\alpha_t}.
\end{align}
Then, DDS update in \eqref{eqn:onestep_dds} can be obtained from the following latent optimization:
\begin{align}
     \min_\z  \| \z - \hat\z_{0|t}(c_\varnothing) \|^2 + {\gamma} R(\z),\quad \mbox{where}\quad R(\z):= \frac{\| \z - \hat\z_{0|t}(c_{tgt}) \|^2}{(1-\gamma)}
\label{eqn:obj_dds}
\end{align}
 Furthermore, it is equivalent to Tweedie's formula with CFG, i.e.:
 \begin{align}\label{eq:dds_CFG}
       \hat\z_{0|t}^\gamma(c_{tgt}) &:= \frac{\z_t - \sqrt{1-\bar\alpha_t}\veps_\theta^\gamma(\z_t, t, c_{tgt})
    }{\sqrt{\bar\alpha_t}} 
\end{align}
where $\veps_\theta^\gamma(\z_t, t, c_{tgt})=\veps_\theta(\z_t, t, c_\varnothing) + \gamma [\veps_\theta(\z_t, t, c_{tgt}) - \veps_\theta(\z_t, t, c_\varnothing)]$.
\end{theorem}
Theorem~\ref{theorem:dds} reveals that the one step latent optimization \eqref{eqn:obj_dds} of DreamSampler reproduces the DDS update \eqref{eqn:onestep_dds}. 
That being said, the main advantages of DreamSampler stems from the added noise to updated source image $\z_0$. 
Specifically, in contrast to the original DDS method that adds newly sampled Gaussian noise to $\z_0$, DreamSampler adds the estimated noise by $\veps_\theta$ in the previous timestep of reverse sampling.
Initiated from the inverted noise, reverse sampling do not deviate significantly from the reconstruction trajectory even though source prompt is not given, because the latent optimization \eqref{eqn:obj_dds} represents a proximal problem that regulates the sampling process.
Consequently, DreamSampler only requires text prompt that describe the editing direction for the real imaging editing.

Finally, it is noteworthy that the equivalent interpretation \eqref{eq:dds_CFG} 
could be readily extended to spatially localized distillation by computing the solution as
\begin{align}
        \hat\z_{0|t}^\gamma=   \left({\z_t - \sqrt{1-\bar\alpha_t}\sum_i { \Mc_i \odot \veps_\theta^\gamma(\z_t, t, c_{tgt}^{(i)})} }\right)/{\sqrt{\bar\alpha_t}}
    \label{eqn:local_dist}
\end{align}
where $\Mc_i$ denotes the pixel-wise mask, $\odot$ is element-wise multiplication,
and $c_{tgt}^{(i)}$ denote $i$th mask and corresponding text prompt pair.
The entire process for the real image editing via DreamSampler is described in Algorithm~\ref{alg:caseB}.
Remark that the Algorithm~\ref{alg:caseB} is equivalent to the Algorithm~\ref{alg:caseA} through Theorem~\ref{theorem:dds}, by setting $\eta \beta_t = 0$ to achieve the deterministic sampling, and by setting $\omega=0$ to ensure that the latent optimization problem is solved during unconditional sampling. Line 9 of the Algorithm~\ref{alg:caseA} is disappeared since we are assuming that $\psi=\z_0$ and $g$ as identity mapping.

\vspace{-0.3cm}
\begin{algorithm}[b]
\setstretch{1.}
\caption{DreamSampler for Image Editing}\label{alg:caseB}
\begin{algorithmic}
\Require source image $\x$, image encoder $\Ec_\phi$, latent diffusion model $\veps_\theta$, null-text embedding $c_\varnothing$, conditioning text embedding $c_{tgt}$.
\State $\z_0 \gets \Ec_\phi(\x)$
\State $\z_T \gets \text{Inversion}(\z_0)$
\For{$t\in [T, 0]$}
\State $\hat\veps_\theta \gets \veps_\theta(\z_t, t, c_\varnothing) + \gamma [\veps_\theta(\z_t, t, c_{tgt}) - \veps_\theta(\z_t, t, c_\varnothing)]$
\State $\bar\z \gets  (\z_t - \sqrt{1-\bar\alpha_t}\hat\veps_\theta)/\sqrt{\bar\alpha_t}$
\State $\z_t \gets \sqrt{\bar\alpha_{t-1}} \bar\z + \sqrt{1-\bar\alpha_{t-1}} \veps_\theta(\z_t, t, c_\varnothing)$
\EndFor
\end{algorithmic}
\end{algorithm}

\subsection{DreamSampler for Inverse Problems}
\vspace{-0.1cm}
DreamSampler can leverage multiple regularization terms to precisely constrain the sampling process to solve inverse problems.
For example, we can solve the text-guided image inpainting task by defining the regularization function as
\begin{align}
    \Rc(\z) = (1-\gamma)\| \Mc \odot( \z - \hat\z_{0|t}(c_{tgt})) \|^2 + \gamma \|\y-\Ac\Dc_\varphi(\z)\|^2, \label{eqn:inverse_reg}
\end{align}
where $\Mc$ denotes the operation to create the measurement by masking out the target region. 
This regularization term implies that the sample image satisfies data consistency for regions where the true signal is preserved, while guiding the masked region to reflect the target text prompt.
When we solve the entire latent optimization problem, we follows two-step approach of TReg~\cite{kim2023regularization}.
For the details, refer to the appendix. 
The main difference with TReg is that we separate the text-guidance from the data consistency term.
In other words, we initialize the $\z$ for the data consistency term as $\hat\z_{0|t}(c_\varnothing)$ while TReg uses $(1-\omega) \hat\z_{0|t}(c_\varnothing) + \omega \hat\z_{0|t}(c_{tgt})$ where the $\omega$ denotes the CFG scale.
This difference allows DreamSampler to solve the inpainting problem with different text-guidance for each masked region, by combining localized distillation approach introduced in Section~\ref{sec:method_caseB}.

\section{Experimental Results}
\label{sec:results}
\subsection{Image Restoration through Vectorization}

\begin{figure}[t]
    \centering
     \includegraphics[width=\linewidth]{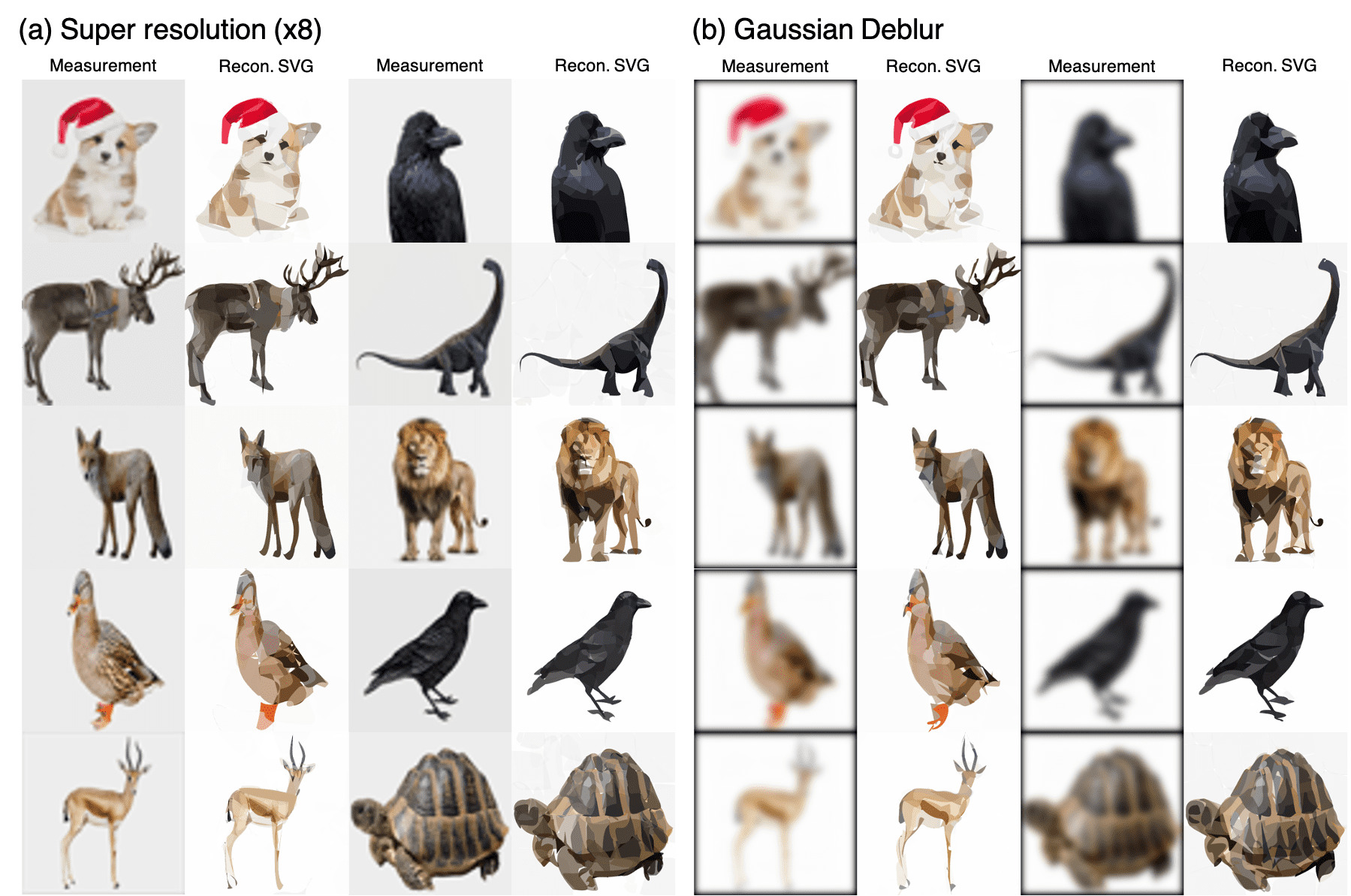}
    \caption{Representative results for image vectorization task with image reconstruction.}
    \vspace{-0.3cm}
    \label{fig:svg_result}
\end{figure}

\begin{figure}[t]
    \centering
    \includegraphics[width=0.7\linewidth]{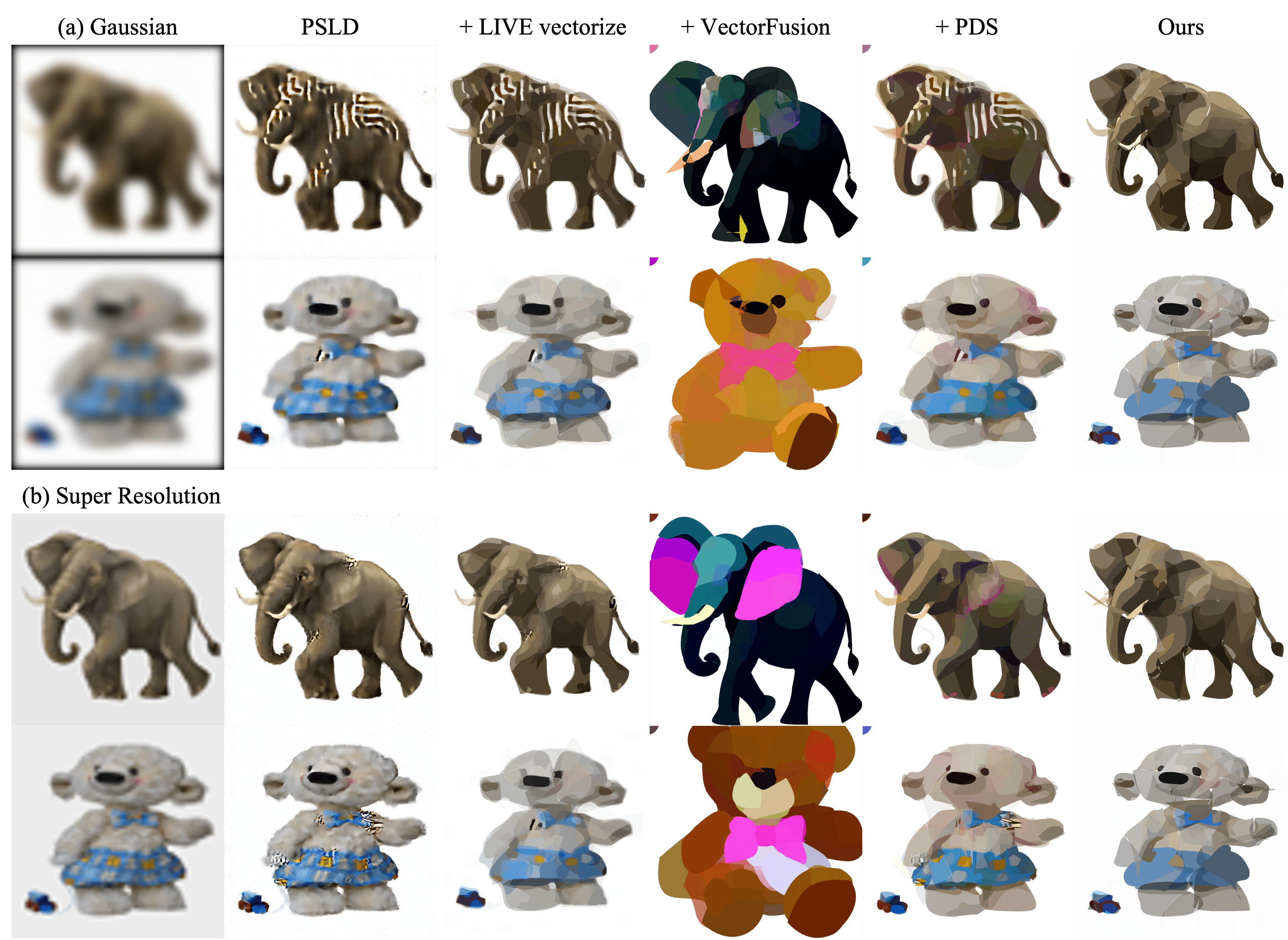}
    \caption{Qualitative comparison of SVG reconstruction. For baselines, we first obtain an initial reconstruction using PSLD \cite{rout2023solving}, vectorize it with LIVE \cite{ma2022towards}, and refine the output vector with VectorFusion \cite{jain2023vectorfusion} or PDS \cite{koo2024posterior}. DreamSampler outperforms this multi-step approach by simultaneously solving the inverse problem and updating SVG parameters via score distillation.
    }
    \vspace{-0.3cm}
    \label{fig:svg_compare}
\end{figure}

As a novel application that other methods have not explored, 
here we present an image vectorization from blurry measurement using DreamSampler.  In this scenario, the generator $g(\cdot)$ corresponds to a differentiable rasterizer (DiffVG \cite{li2020differentiable}), and the parameters $\psi$ of the generator consist of path parameters comprising Scalable Vector Graphics (SVG).  

Specifically, we address a text-guided SVG inverse problem, acknowledging that low-quality, noisy measurements can detract from the detail and aesthetic quality of vector designs. 
Here, our goal is to accurately reconstruct SVG paths for the parameter $\psi$ that, when rasterized, align closely with the provided measurements $\y$ and text conditions $c_{\mathbf{y}}$, related through the forward measurement operator $\Ac$. In this context, the regularizer $\Rc$ in \eqref{eqn:our_optim} corresponds to the data consistency term. 
Forward measurement operators are specified as follows: (\textbf{a}) For super-resolution, bicubic downsampling is performed with scale $\times8$. (\textbf{b}) For Gaussian blur, the kernel has size $61 \times 61$ with a standard deviation of 5.
Then, the latent optimization framework of the SVG inverse problem is defined as:
\begin{equation}
    \min_\psi (1 - \gamma)\lambda_{SDS} \| \Ec (g(\psi)) - \hat\z_{0|t}(c_{\y}) \|^2 + \gamma\lambda_{DC} \|\y-\Ac g(\psi)\|^2,
    \label{eqn:svg_optim}
\end{equation}

where we found out that $\gamma = \bar \alpha_t$ works well in practice. SVG primitives $\psi$ are initialized with radius 20, random fill color, and opacity uniformly sampled between 0.7 and 1, following \cite{jain2023vectorfusion}.  In this paper, we use closed Bézier curves for iconography artistic style. For the optimization of \eqref{eqn:svg_optim}, we use Adam optimizer with ($\beta_1, \beta_2$)=(0.9, 0.9). For the text condition $c_{\y}$, we append a suffix to the object\footnote{e.g. "\{\textit{a cute fox}\}, \textit{minimal flat 2d vector icon. lineal color. on a white background. trending on artstation}."}. Additional experimental details are provided in the appendix.

Comparatively, the proposed framework is evaluated against a multi-stage baseline approach involving initial rasterized image reconstruction using the state-of-the-art solver (PSLD \cite{rout2024solving}) that is based on the latent diffusion model. Then, we vectorize the reconstruction using the off-the-shelf Layer-wise Image Vectorization program (LIVE \cite{ma2022towards}). An optional step includes refining the SVG output via latent score distillation sampling algorithms, exemplified by VectorFusion \cite{jain2023vectorfusion} and Posterior Distillation Sampling (PDS, \cite{koo2024posterior}).

Figure \ref{fig:svg_result} illustrates that the proposed framework achieves high-quality SVG reconstructions with semantic alignment closely matching the specified text condition $c_{\mathbf{y}}$. In contrast, Figure \ref{fig:svg_compare} highlights the deficiencies of the multi-stage baseline methods, particularly its inability to retain detailed fidelity. Errors accumulate during the initial restoration process, resulting in blurriness and undesirable path overlap in the vector outputs. While the solver may achieve adequate reconstructions, the subsequent vectorization step disregards text caption $c_{\y}$, leading to the potential loss of details and semantic coherence. Additional VectorFusion fine-tuning loses consistency with the measurement. Conversely, DreamSampler effectively restores SVGs by directly utilizing latent-space diffusion and text conditions for both vectorization and reconstruction, ensuring the preservation of detail and contextual relevance.

\subsection{Real Image Editing}
\begin{figure}[t]
    \centering
    \includegraphics[width=0.85\linewidth]{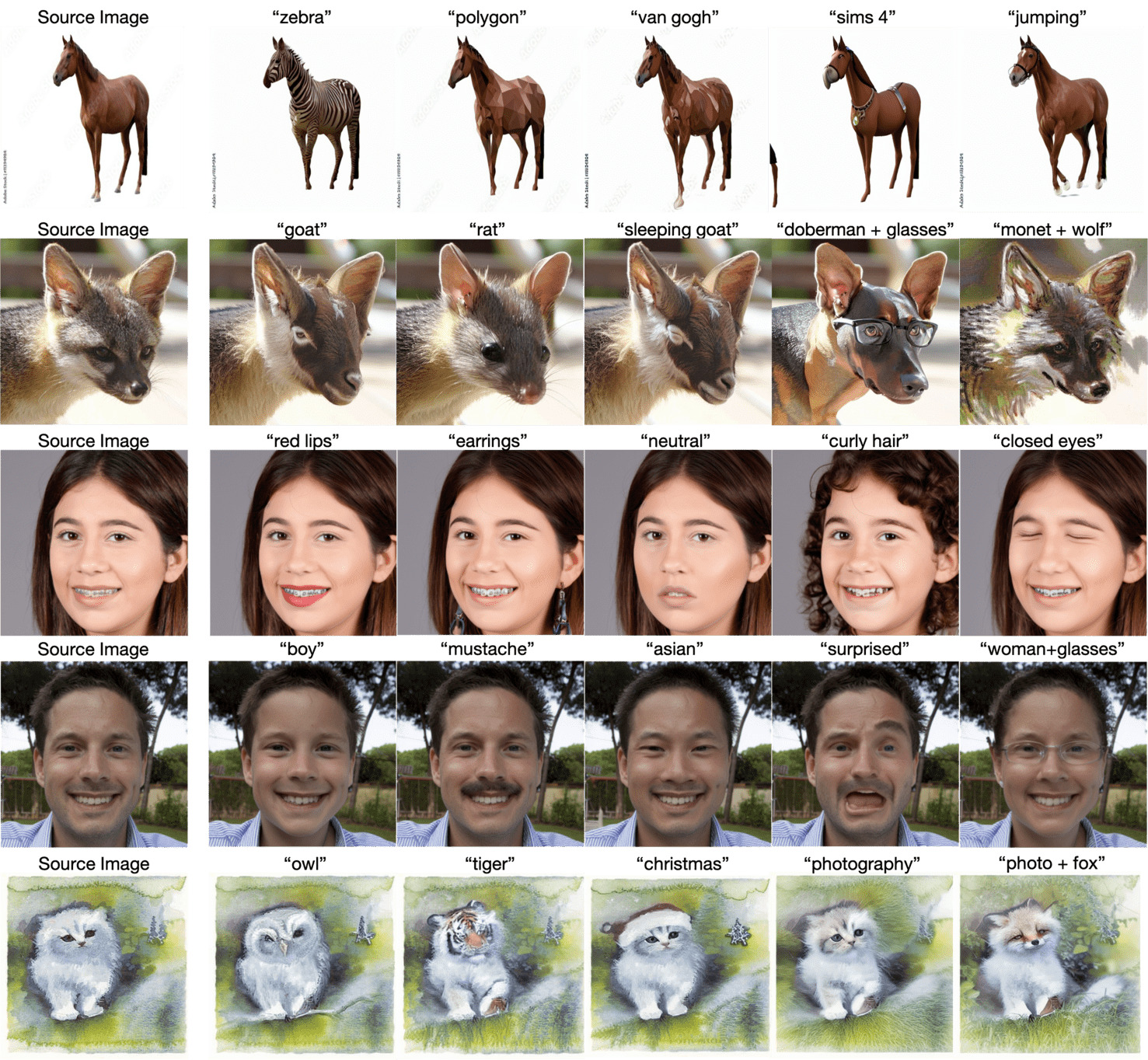}
    \caption{Representative results for real image editing via distillation through reverse sampling. We leverage source images from various domains and the caption above each image denotes text prompt reflecting the editing direction. The result demonstrates that distillation could be effectively conducted during the reverse sampling.}
    \vspace{-0.3cm}
    \label{fig:dds_result}
\end{figure}
For real image editing via DreamSampler, we leverage the Stable-Diffusion v1.5 provided by HuggingFace.
We use linear time schedule and set NFE to 200 for both the DDIM inversion and reverse sampling. For more details on hyper-parameter setting, please refer to the appendix.

To demonstrate the ability of DreamSampler in real image editing, we leverage source images from various domains, including photographs of animals, human faces and drawings.
All results in Figure~\ref{fig:dds_result} show that DreamSampler accurately reflects the provided text prompts for editing.
Specifically, DreamSampler does not change bias components, which are intended not to be edited by text prompt. For instance, in 3rd and 4th rows of Figure~\ref{fig:dds_result}, features such as braces and background including striped shirt, are well-maintained while the text prompts are accurately reflected. 
Moreover, DreamSampler effectively reflects multiple editing directions simultaneously such as "doberman" $+$ "wearing glasses", "woman" $+$ "wearing glasses" or "photography" $+$ "fox".
\begin{figure}[t]
    \centering
    \includegraphics[width=0.75\linewidth]{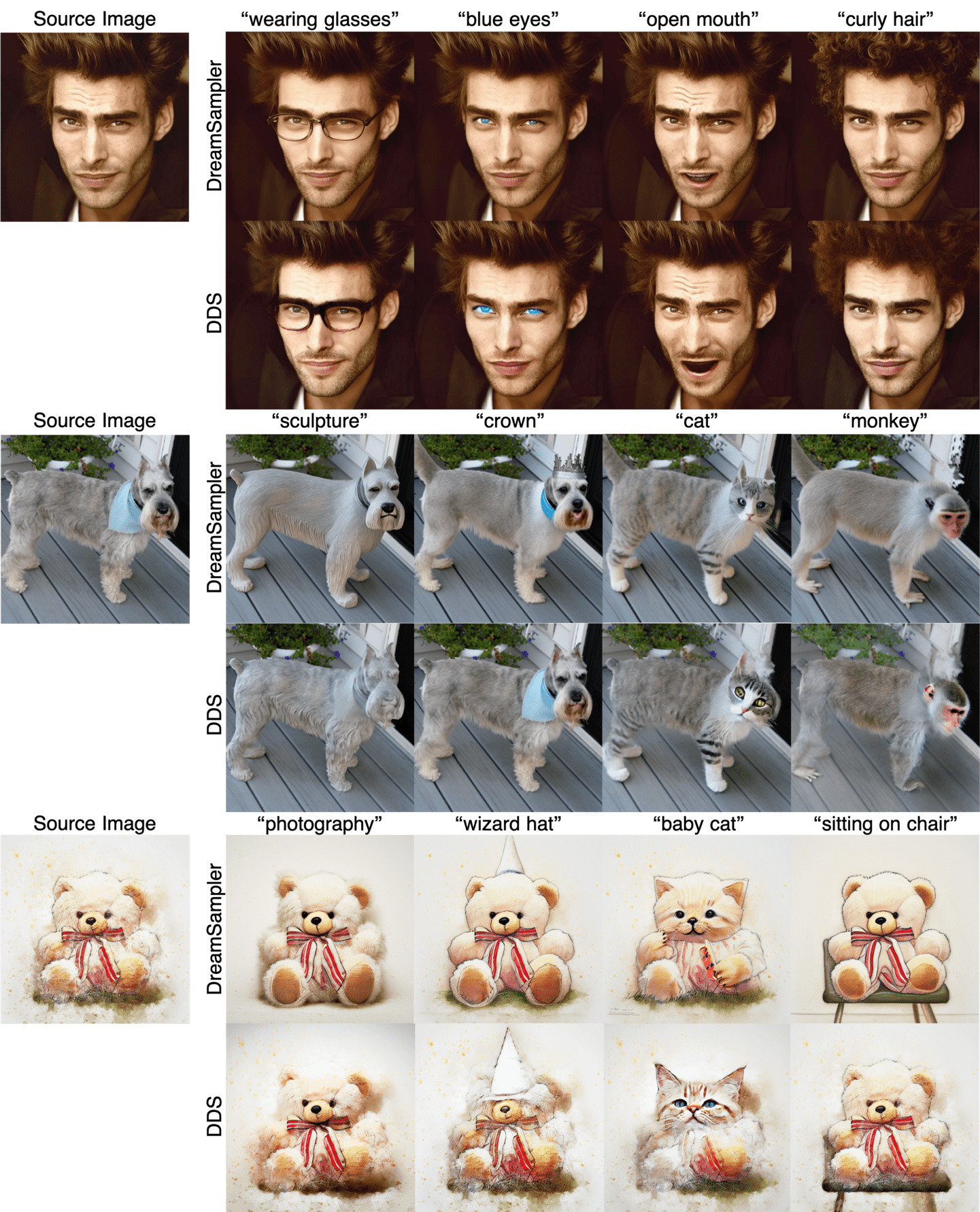}
    \caption{Qualitative comparison with DDS for the real image editing. Both DreamSampler and DDS edit images following target text, but DreamSampler achieves higher fidelity with preserving bias component.}
    \label{fig:dds_compare}
    \vspace{-0.3cm}
\end{figure}
We next conduct a qualitative comparison with the original DDS algorithm to show the improvement from integration of distillation into reverse sampling. As illustrated in Figure~\ref{fig:dds_compare}, DreamSampler is capable of editing according to text prompt robustly across various domains of source images.
Furthermore, DreamSampler achieves better fidelity and bias component preservation compared to the original DDS.

For the quantitative comparison, we compare DreamSampler against diffusion-based editing algorithms~\cite{meng2021sdedit, hertz2022prompt, mokady2023null}, following the experimental setups of \cite{park2024energy, parmar2023zero}. We use the prompt "a photograph of \{\}" with the source and target objects inserted. For the CFG scale, we use $0.15 \bar\alpha_t$ across all cases. Table~\ref{tab:edit} demonstrates that DreamSampler outperforms most baselines in image editing tasks. Specifically, DreamSampler generates glasses more naturally in the "cat $\rightarrow$ cat w/ glasses" task, improves fidelity in source image edits compared to baseline algorithms as shown in Figure~\ref{fig:edit_c2cg}. Some baselines achieve high CLIP accuracy by focusing on the generation of glasses, regardless of its natural appearance.

\begin{table}[t]
\caption{Comparison to diffusion-based editing methods. Dist for DINO-ViT Structure Distance. Baseline results are from~\cite{park2024energy}.}
\centering
\resizebox{0.7\linewidth}{!}{
\begin{tabular}{ccccccc}
\toprule
\multirow{2}{*}{\textbf{Method}} & \multicolumn{2}{c}{\textbf{Cat$\rightarrow$Dog}} & \multicolumn{2}{c}{\textbf{Horse$\rightarrow$Zebra}} & \multicolumn{2}{c}{\textbf{Cat$\rightarrow$ Cat w/ glasses}} \\
& CLIP-Acc $\uparrow$ & Dist $\downarrow$& CLIP-Acc $\uparrow$& Dist $\downarrow$& CLIP-Acc $\uparrow$& Dist $\downarrow$\\
\midrule
SDEdit + word swap 
& 71.2\% & 0.081 & 92.2\% & 0.105 & 34.0\% & 0.082 \\
DDIM + word swap
& 72.0\% & 0.087 & 94.0\% & 0.123 & 37.6\% & 0.085 \\
prompt to prompt 
& 66.0\% & 0.080 & 18.4\% & 0.095 & 69.6\% & 0.081  \\
p2p-zero
& 92.4\% & 0.044 & 75.2\% & 0.066 & 71.2\% & 0.028  \\
EBCA
& 93.7\% & 0.040 & 90.4\% & 0.061 & 81.1\% & 0.052  \\
\color{teal} DreamSampler (Ours)
& 90.3\%&  \textbf{0.029}&  \textbf{95.2}\% &  \textbf{0.038}&  48.3\% &  \textbf{0.025}  \\
\bottomrule
\end{tabular}
}
\label{tab:edit}
\end{table}

\begin{table}[t]
\caption{Comparison to text-conditioned diffusion-based inpainting solvers. \textbf{Bold}: the best score, \underline{Underline}: the second best.}
\centering
\resizebox{0.7\linewidth}{!}{
\begin{tabular}{ccccccc}
\toprule
\multirow{2}{*}{Method} & \multicolumn{3}{c}{\textbf{Glasses}} & \multicolumn{3}{c}{\textbf{Smile}} \\
& PSNR $\uparrow$ & FID $\downarrow$& CLIP-sim  $\uparrow$& PSNR  $\uparrow$& FID $\downarrow$& CLIP-sim  $\uparrow$\\
\midrule
Stable-Inpaint
& 19.82 & \underline{54.26} & \underline{0.281} & 25.33 & \textbf{19.22} & \textbf{0.249} \\
\midrule
TReg 
& \underline{21.97} & 61.04 & \textbf{0.288} & \underline{26.71} & 24.48 & \textbf{0.249} \\
\color{teal}DreamSampler (Ours)
& \textbf{24.61} & \textbf{27.10} & 0.263 & \textbf{27.90} & \underline{24.33} & \underline{0.242} \\
\bottomrule
\end{tabular}
}
\label{tab:inpaint}
\end{table}

\subsection{Text-guided Image Inpainting}
\begin{figure}[t]
    \centering
    \includegraphics[width=0.9\linewidth]{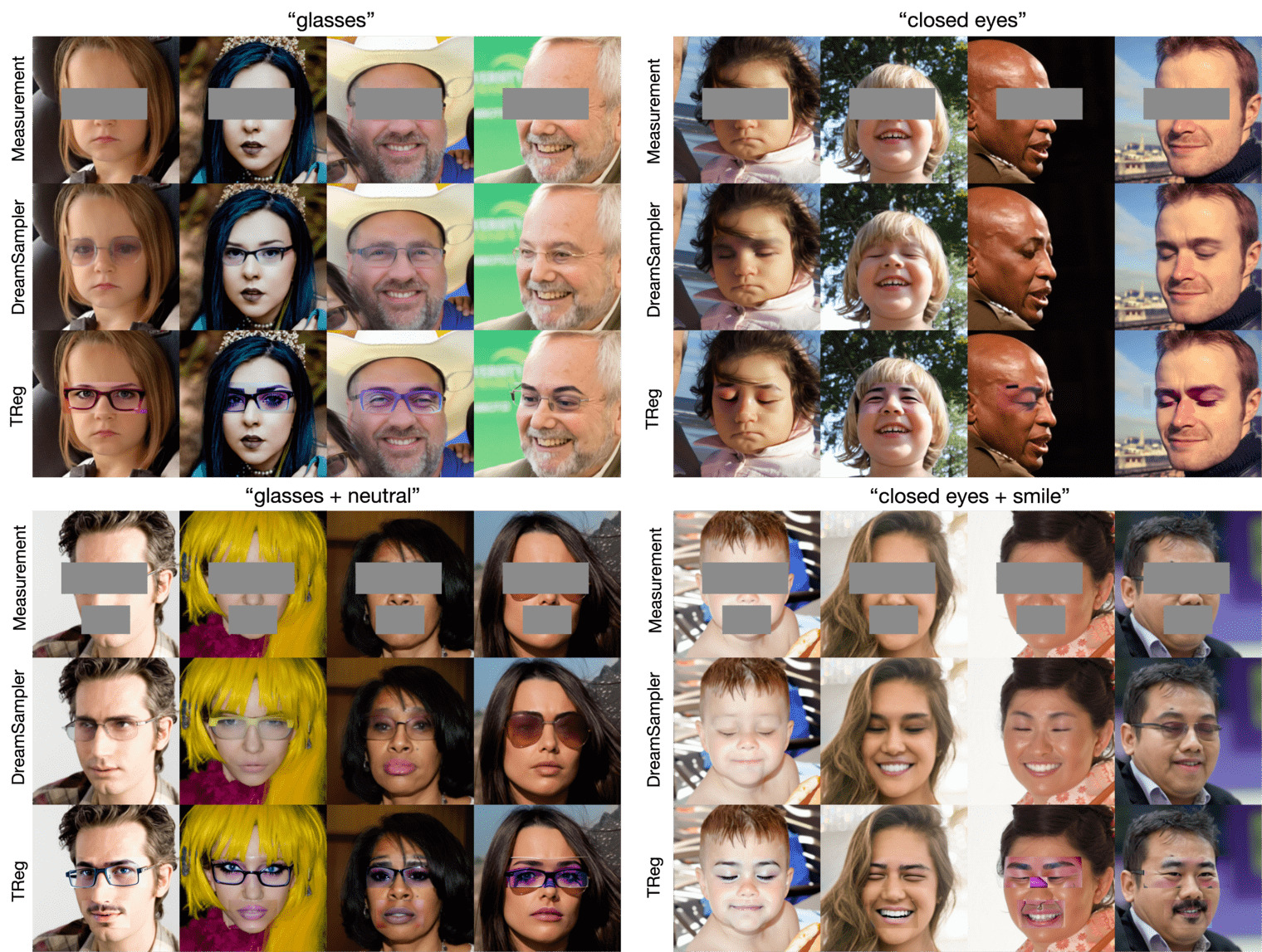}
    \caption{Qualitative comparison for text guided image inpainting task. DreamSampler can generate more realistic images according to given text prompt.}
    \label{fig:inpaint}
    \vspace{-0.3cm}
\end{figure}

For the text-guided image inpainting task, we also use Stable-Diffusion v1.5, linear time schedule, and 200 NFE. In addition to solving \eqref{eqn:inverse_reg}, we apply DPS~\cite{chung2022diffusion} steps during sampling by following TReg~\cite{kim2023regularization} to enhance the consistency of masked region and other regions.
We generate measurements by masking out two rectangular regions on the eyes and mouth, where the masked region is determined based on the averaged face of the 1k FFHQ validation set. The size of measurement is $512\times512$. We solve the inpainting problem by giving text prompts "a photography of face wearing glasses" and "a photography of face with smile".
Figure~\ref{fig:inpaint} shows that DreamSampler fills masked region by reflecting given text prompt accurately. 
While the text guidance is applied inside the mask, data consistency gradients combined with the reconstruction by inversion is applied outside the mask, which results in superior fidelity of the output image. Note that we display the image output as is without any post-processing such as projection\footnote{For linear operator $\Ac$ and measurement $\y$, the projection means $\Ac^\top\Ac \y + (\rmI-\Ac^\top\Ac)\x$.}.
The bottom row of Figure~\ref{fig:inpaint} depict the solution for inpainting problem with two different masks and distinct text guidance. Through localized distillation gradient, DreamSampler generates solutions according to the provided guidance.
DreamSampler generates masked regions with better robustness than TReg. Additionally, in the case of multiple masks, DreamSampler successfully reflects text in the correct regions via localized distillation gradients, whereas TReg fails to meet one of the conditions. For more results, refer to appendix.

We also evaluate both the quality of reconstructions (PSNR, FID) and the accuracy of the text guidance (CLIP similarity) on 1k FFHQ validation set. 
For baselines, we select diffusion-based image inpainting models~\cite{rombach2022high, lugmayr2022repaint} and a text-guided inverse problem solver~\cite{kim2023regularization}. Specifically, Stable-Inpaint~\cite{rombach2022high} is a fine-tuned StableDiffusion model specialized for inpainting task.
Table~\ref{tab:inpaint} shows that DreamSampler outperforms the baselines in terms of PSNR and FID score while achieving comparable CLIP similarity. Stable-Inpaint achieves a lower FID score, while DreamSampler, serving as a zero-shot inpainting solver, achieves a higher PSNR. This suggests superior reconstruction quality of Dreamsampler through data consistency update.

\section{Conclusion}
\label{sec:conclude}
\vspace{-0.2cm}
We presented  DreamSampler, a unified framework of reverse sampling and score distillation by taking the advantages of each algorithms. Specifically, we connected two distinct algorithms under the perspective of latent optimization problem. Consequently, we introduced a generalized optimization framework, which offers new design space to solve various applications. Especially, DreamSampler enables to combine various regularization functions to constrain the sampling process.
Additionally, we provided three applications including image vectorization with reconstruction, real image editing, and image inpainting. DreamSampler could be extended to other algorithms by defining appropriate regularization functions.
The codebase is available to public at \url{https://github.com/DreamSampler/dream-sampler}.

\noindent\textbf{Potential negative social impact. }
The performance of algorithms established on DreamSampler heavily depends on the prior of diffusion model. Hence, the proposed method basically influenced by the potential negative impacts of LDM itself. Thus, proper political regulation is required to mitigate these risks.

\subsubsection*{Acknowledgments}
This work was supported by the National Research Foundation of Korea under Grant RS-2024-00336454, by Institute of Information \& communications Technology Planning \& Evaluation (IITP) grant funded by the Korea government(MSIT)) (No. RS-2022-II220984, Development of Artificial Intelligence Technology for Personalized Plug-and-Play Explanation and Verification of Explanation, and No.2019-0-00075, Artificial Intelligence Graduate School Program(KAIST)), by Culture, Sports and Tourism R\&D Program through the Korea Creative Content Agency grant funded by the Ministry of Culture, Sports and Tourism in 2023, by Field-oriented Technology Development Project for Customs Administration funded by the Korea government (the Ministry of Science \& ICT and the Korea Customs Service) through the National Research Foundation (NRF) of Korea under Grant NRF2021M3I1A1097910.


%
%
\bibliographystyle{splncs04}
\bibliography{main}

\renewcommand{\thefigure}{S\arabic{figure}}
\renewcommand{\thetable}{S\arabic{table}}

\title{Supplementary material of "DreamSampler: Unifying Diffusion Sampling and Score Distillation for Image Manipulation"} 

\titlerunning{DreamSampler}
\authorrunning{Kim et al.}
\author{Jeongsol Kim\inst{1, *}\and
Geon Yeong Park\inst{1, *} \and
Jong Chul Ye\inst{2}}



\institute{Dept. of Bio \& Brain Engineering, KAIST\and
Kim Jae Chul AI graduate school, KAIST\\
\email{\{jeongsol, pky3436, jong.ye\}@kaist.ac.kr}\\
* Equal contribution}

\maketitle

\section{Proof of Theorem 1.}
The solution of latent optimization problem
\begin{align}
    \bar\z = \argmin_\z \| \z - \hat\z_{0|t}(c_\varnothing) \|^2 + \frac{\gamma}{1-\gamma} \| \z - \hat\z_{0|t}(c_{ref}) \|^2
\end{align}
is computed as a closed form of
\begin{align}
    \bar\z &= (1-\gamma) \hat\z_{0|t}(c_\phi) + \gamma \hat\z_{0|t}(c_{ref})
\end{align}
because the objective function is convex and
\begin{align}
    2(1-\gamma) (\bar\z - \hat\z_{0|t}(c_\phi)) + 2\gamma (\bar\z -\hat\z_{0|t}(c_{ref})) = 0.
\end{align}
Then,
\begin{align}
    \bar\z &= (1-\gamma) \hat\z_{0|t}(c_\varnothing) + \gamma \hat\z_{0|t}(c_{ref})\\
    &= \hat\z_{0|t}(c_\varnothing) - \gamma [\hat\z_{0|t}(c_\varnothing)- \hat\z_{0|t}(c_{ref})]\\
    &= \hat\z_{0|t}(c_\varnothing) - \gamma \left[\frac{\z_t-\sqrt{1-\bar\alpha_t}\veps_\theta(\z_t, c_\varnothing)}{\sqrt{\bar\alpha_t}} - \frac{\z_t - \sqrt{1-\bar\alpha_t}\veps_\theta(\z_t, c_{ref})}{\sqrt{\bar\alpha_t}}\right]\\
    & = \hat\z_{0|t}(c_\varnothing) - \gamma \frac{\sqrt{1-\bar\alpha_t}}{\sqrt{\bar\alpha_t}}[\veps_\theta(\z_t, c_\varnothing)-\veps_\theta(\z_t, c_{ref})]
\end{align}
Finally, by substituting Tweedie's formula into $\hat\z_{0|t}(c_\varnothing)$, we can get
\begin{align}
    \bar\z &= \frac{\z_t - \sqrt{1-\bar\alpha_t}[\veps_\theta(\z_t, c_\varnothing)+\gamma\{\veps_\theta(\z_t, c_{ref}) - \veps_\theta(\z_t, c_\varnothing)\}]}{\sqrt{\bar\alpha_t}} 
\end{align}
This concludes the proof.

\section{Implementation details}
In this section, we provide details on the implementation and experimental settings of DreamSampler. 

\subsection{Image Restoration through Vectorization}
For the reverse sampling, we set the CFG scale to 100, and NFE to 1000 for both super-resolution and Gaussian deblurring. For the Lagrangian coefficient, we set $\lambda_{SDS}=2.4, \lambda_{DC}=3$ for Guassian deblur and $\lambda_{SDS}=1, \lambda_{DC}=4$ for super-resolution. Other optimization configurations, e.g. learning rate, optimization algorithms, paths, etc, follow \cite{jain2023vectorfusion}. Specifically, a self-intersection regularizer $\mathcal{L}_{xing}$ is used with a weight $0.01$. 
The learning rate initiates at 0.02 and linearly escalates to 0.2 across 500 steps, subsequently undergoing a cosine decay to 0.05 upon optimization completion for control point coordinates. Fill colors are subjected to a learning rate that is 20 times lower than that of control points, while the solid background color is allocated a learning rate 200 times lower.

\subsection{Real Image Editing}

We utilize the DDIM inversion using the null-text to initialize the latent $\z_T$. The null-text is defined by empty text "" for both inversion and sampling, which could be leveraged in general.
For the noise addition steps at each timestep, we leverage the deterministic sampling by setting $\eta\beta_t=0$.
For the CFG scale $\gamma$, we found that time-dependent value between $0.1\bar\alpha_t$ and $0.3\bar\alpha_t$ can edit image with robustness.
As $\bar\alpha_t \rightarrow 1$ when $t\rightarrow 0$, this scale allows early-stage sampling to reconstruct the source image while the later-stage sampling reflects guided direction, according to \eqref{eqn:obj_dds}.
Here, $\gamma$ indicates the interpolation coefficient for the latent optimization problem
\begin{equation}
    \label{eqn:dds_obj}
    \min_\z (1-\gamma) \| \z - \hat\z_{0|t}(c_\phi) \|^2 + \gamma \| \z - \hat\z_{0|t}(c_{tgt})\|^2,
\end{equation}
which is analyzed in section~\ref{sec:gamma_edit}.

\subsection{Text-guided Image Inpainting}

For the inapinting task, we utilize the DDIM inversion with null-text where the null-text is "out of focus, depth of field" to adopt concept negation for better generation quality.
After solving the latent optimization problem, we add stochastic noise by setting $\eta\beta_t=\sqrt{\bar\alpha_t}\sqrt{1-\bar\alpha_{t-1}}$.
As other tasks, we set NFE to 200.

From the general formulation of DreamSampler, we can derive the latent optimization problem for the inpainting task.
Suppose that the generator $g=\Dc_\varphi$ and $\phi=\z$ so the generator maps the latent vector to pixel space. Note that $\Dc_\varphi$ is a component of autoencoder that satisfies $\z= \Ec_\phi(\Dc_\varphi(\z))$ called perfect reconstruction constraint.
Then, by defining the regularization function as
\begin{align}
    \Rc(\Dc_\varphi(\z)) = \|\z - \hat\z_{0|t}(c_{tgt})\|^2 + \|\y-\Ac\Dc_\varphi(\z) \|^2
\end{align}
we can reach to the latent optimization problem of
\begin{align}
    \min_{\z, \x}  \gamma_1 \underbrace{\| \z - \hat\z_{0|t}(c_{tgt}) \|^2}_{\text{Score distillation}} &+ \gamma_2 \underbrace{\| \z - \hat\z_{0|t}(c_\phi) \|^2}_{\text{Proximal term}}\nonumber\\
    &+ \gamma_3 \underbrace{\left[ \| \y - \Ac \Dc_\varphi(\z) \|^2 + \|\z - \Ec_\phi(\x) \|^2 \right]}_{\text{Data consistency}}
\end{align}
where the last term comes from the perfect reconstruction constraint, and $\gamma_1 + \gamma_2 + \gamma_3 = 1$.
The objective function comprises three components: data consistency, which ensures alignment between the current estimate and the given measurement; score distillation, which serves as textual guidance; and a proximal term, which regulates the solution to remain close to the inverted trajectory.
Following TReg~\cite{kim2023regularization}, we solve the optimization problem sequentially. First, using the approximation $\x = \Dc_\varphi(\z)$, the optimization problem with respect to $\x$ becomes 
\begin{align}
    \min_x \| \y - \Ac\x \|^2 + \|\z - \Ec_\phi(\x) \|^2 + \lambda \| \x - \Dc_\phi(\z) + \etab \|^2
\end{align}
where the dual variable $\etab$ is set to a zero vector for simplicity.
Then, by initializing $\z=\hat\z_{0|t}(c_\phi)$, we have
\begin{align}
    \hat\x_0(\y) = \argmin_\x \| \y - \Ac(\x) \|^2 + \lambda \| \x - \Dc_\varphi(\hat\z_{0|t}(c_\phi)) \|^2.
\end{align}
This could be solved by conjugate gradient (CG) method.
Subsequently, using the approximation $\z = \Ec_\phi(\x)$ with $\etab=\mathbf{0}$, the optimization with respect to $\z$ becomes
\begin{align}
    \min_z \gamma_1 \| \z - \hat\z_{0|t}(c_{tgt})\|^2 + \gamma_2 \|\z - \hat\z_{0|t}(c_\phi) \|^2 + \gamma_3 \|\z- \Ec_\phi(\hat\x_0(\y)) \|^2,
\end{align}
which leads to a closed-form solution
\begin{align}
    \bar\z = \bar\alpha_t\hat\z_{0|t}(c_{tgt}) + (1-\bar\alpha_t)^2 \hat\z_{0|t}(c_\phi) + \bar\alpha_t(1-\bar\alpha_t)\hat\z_{0|t}(\y)
\end{align}
where $\hat\z_{0|t}(\y):=\Ec_\phi(\hat\x_0(\y))$.
Specifically, we apply localized distillation gradient by leveraging a mask so the objective function inside/outside of masked region is differ as 
\begin{align}
    \min_z \gamma_1 \| \Mc \odot (\z - \hat\z_{0|t}(c_{tgt}))\|^2 + \gamma_2 \|\z - \hat\z_{0|t}(c_\phi) \|^2 + \gamma_3 \|\z- \Ec_\phi(\hat\x_0(\y)) \|^2
\end{align}
where $\Mc$ denotes pixel-wise mask with ones inside the masked region and zeros elsewhere, while $\odot$ denotes element-wise multiplication.
Hence, the closed form solution is described as
\begin{align}
    \bar\z = 
    \begin{cases}
      \bar\alpha_t\hat\z_{0|t}(c_{tgt}) + (1-\bar\alpha_t)^2 \hat\z_{0|t}(c_\phi) + \bar\alpha_t(1-\bar\alpha_t)\hat\z_{0|t}(\y) & \text{inside mask}\\
      (1-\bar\alpha_t) \hat\z_{0|t}(c_\phi) + \bar\alpha_t\hat\z_{0|t}(\y) & \text{otherwise}\\
    \end{cases}
    \label{eqn:closed_solution}
\end{align}

Also, we use additional DPS step to ensure the consistency between inside and outside of masked region. By following TReg, we compute the \eqref{eqn:closed_solution} when $\Gamma = \{t|t \text{ mod } 3 = 0, t \leq 170\}$ and apply DPS gradie   nt otherwise. In summary, the pseudocode of DreamSampler for the inpainting task is described as Algorithm~\ref{alg:inpaint}

\begin{algorithm}[h!]
\setstretch{1.}
\caption{DreamSampler for Image Inpainting}\label{alg:inpaint}
\begin{algorithmic}
\Require measurement $\y$, image encoder $\Ec_\phi$, latent diffusion model $\veps_\theta$, null-text embedding $c_\emptyset$, conditioning text embedding $c_{tgt}$.
\State $\z_0 \gets \Ec_\phi(\y)$
\State $\z_T \gets \text{Inversion}(\z_0)$
\For{$t\in [T, 0]$}
    \State $\hat\veps_\theta(c_\emptyset), \hat\veps_\theta(c_{tgt}) \gets \veps_\theta(\z_t, t, c_\emptyset), \veps_\theta(\z_t, t, c_{tgt})$
    \State $\veps \sim \Nc(0, \rmI)$
    \State {$\tilde\veps_t \gets (\sqrt{1-\bar\alpha_{t-1}-\eta^2\beta_t^2}\hat\veps_\theta(c_\emptyset) + \eta\beta_t\veps)/\sqrt{1-\bar\alpha_{t-1}}$}
    \If{$t\in \Gamma$}
        \State $\hat\z_{0|t}(c_\emptyset) \gets (\z_t - \sqrt{1-\bar\alpha_t}\hat\veps_\theta(c_\emptyset))/\sqrt{\bar\alpha_t}$
        \State $\hat\z_{0|t}(c_{tgt}) \gets (\z_t - \sqrt{1-\bar\alpha_t}\hat\veps_\theta(c_{tgt}))/\sqrt{\bar\alpha_t}$
        \State $\hat\x_0(\y) \gets \argmin_\x \| \y-\Ac(\x) \|^2 + \lambda \| \x - \Dc_\varphi(\hat\z_{0|t}(c_\emptyset))\|^2$
        \State $\bar\z_{in} \gets \bar\alpha_t \hat\z_{0|t}(c_{tgt}) + (1-\bar\alpha_t)^2\hat\z_{0|t}(c_\emptyset) + \bar\alpha_t(1-\bar\alpha_t)\Ec_\phi(\hat\x_0(\y))$
        \State $\bar\z_{out} \gets (1-\bar\alpha_t)\hat\z_{0|t}(c_\emptyset) + \bar\alpha_t\Ec_\phi(\hat\x_0(\y))$
        \State $\bar\z \gets \Mc \odot \bar\z_{in} + (1-\Mc)\odot \bar\z_{out}$
        \State {$\z_{t-1} \gets \sqrt{\bar\alpha_{t-1}}\bar{\z}+ \sqrt{1-\bar\alpha_{t-1}}\tilde\veps_t$}
    \Else
        \State $\hat\z_{0|t}(c_\emptyset) = (\z_t - \sqrt{1-\bar\alpha_t}\hat\veps_\theta(c_\emptyset))/\sqrt{\bar\alpha_t}$
        \State $\z_{t-1}' \gets \sqrt{\bar\alpha_{t-1}}\hat{\z}_{0|t}(c_\emptyset) + \sqrt{1-\bar\alpha_{t-1}}\tilde\veps_t$
        \State $\z_{t-1} \gets \z'_{t-1} - \rho_{t}\nabla_{\z_t}\|\Ac(\Dc_{{\varphi}}(\hat\z_{0|t})) - \y\|$
    \EndIf 
\EndFor
\end{algorithmic}
\end{algorithm}

\section{More qualitative comparison for text-guided inpainting}
\begin{figure}[h!]
    \centering
    \includegraphics[width=\linewidth]{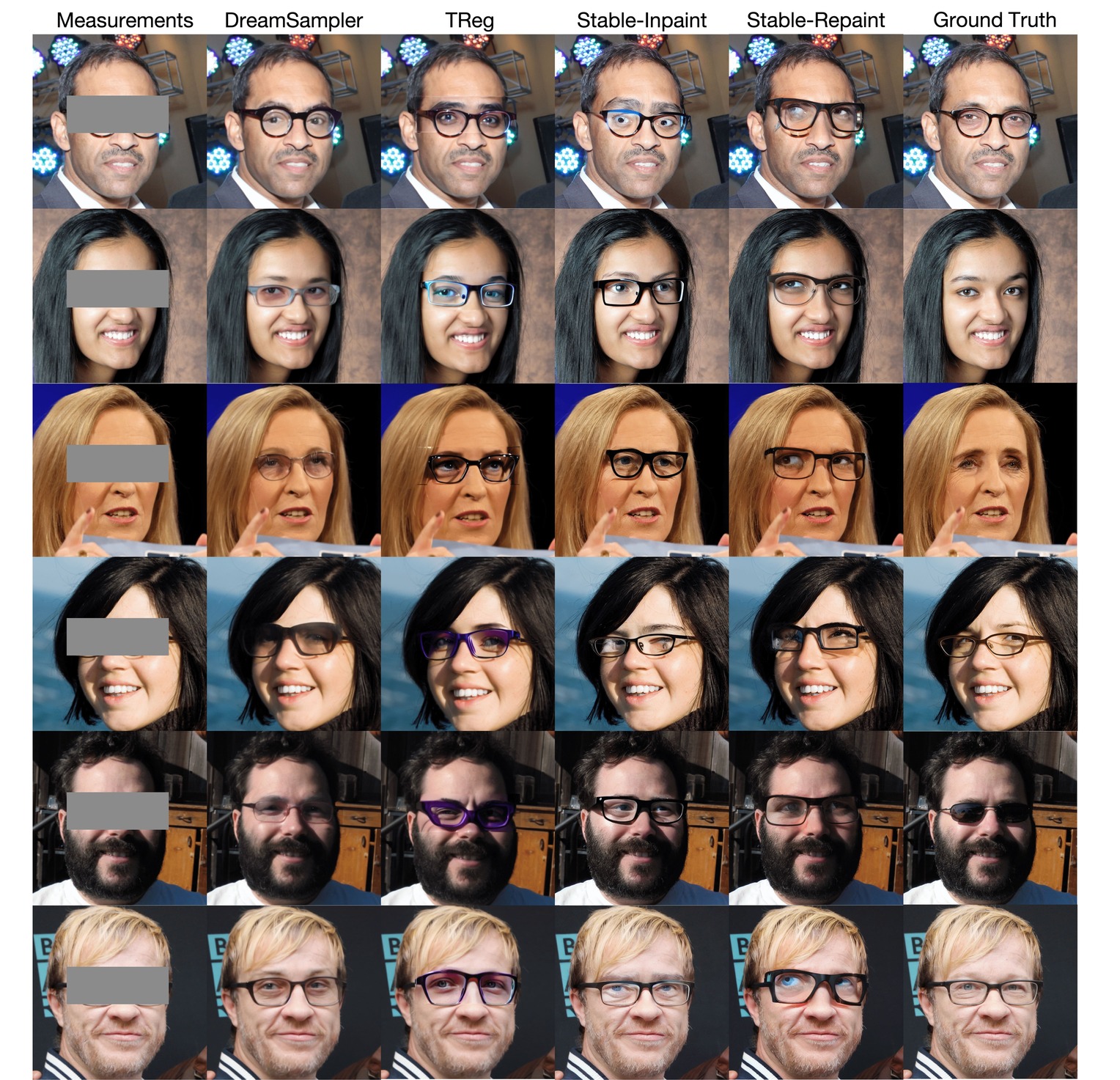}
    \caption{Qualitative comparison for inpainting task with 512x512 FFHQ dataset. Text prompt "A photography of face wearing glasses" is given.}
    \label{fig:inpaint_glass_comp}
\end{figure}

\begin{figure}[h!]
    \centering
    \includegraphics[width=\linewidth]{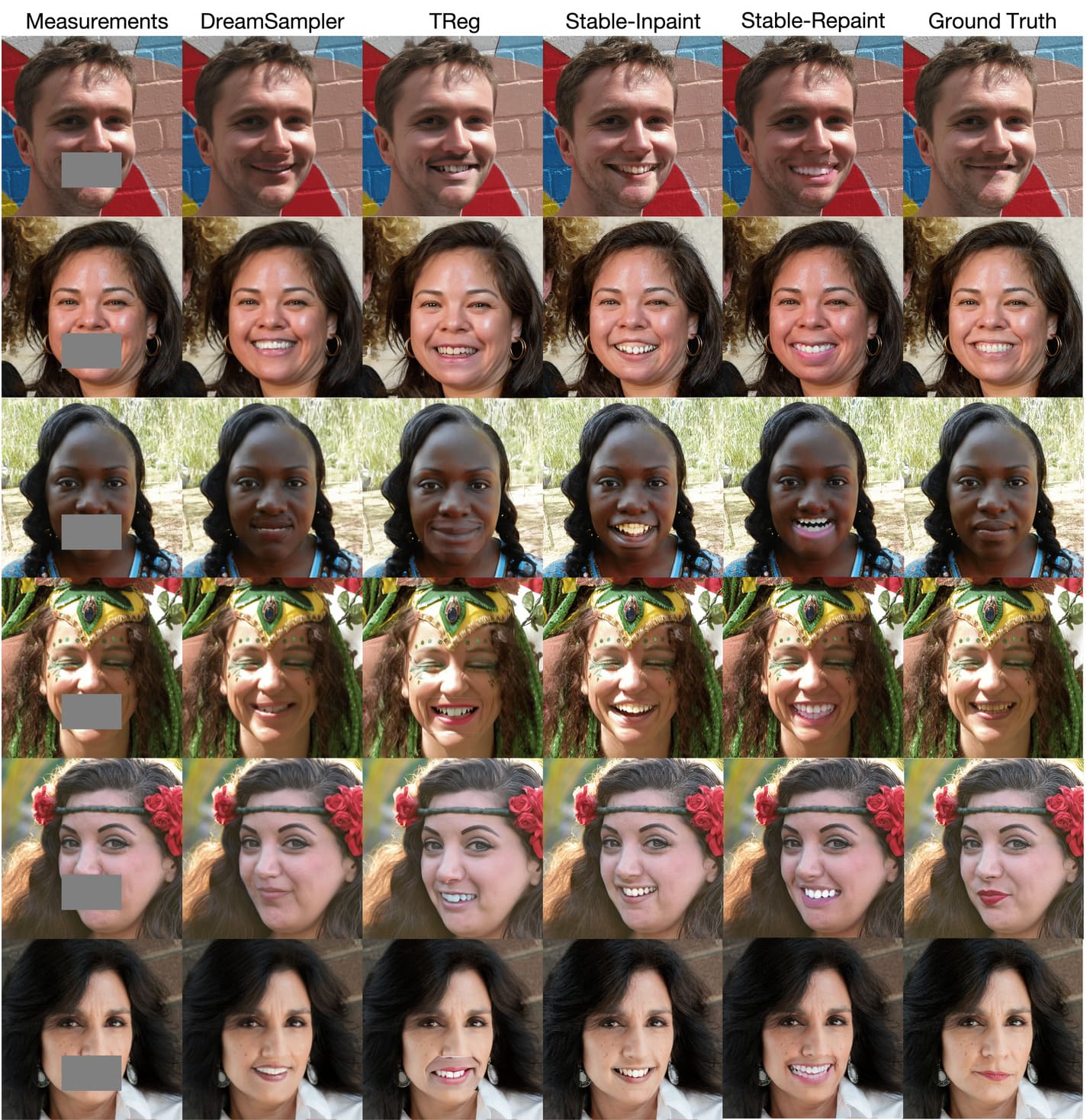}
    \caption{Qualitative comparison for inpainting task with 512x512 FFHQ dataset. Text prompt "A photography of face with smile" is given.}
    \label{fig:inpaint_smile_comp}
\end{figure}

In this section, we show qualitative comparison of the DreamSampler for text-guided image inpainting tasks, against multiple state-of-the-art diffusion-based algorithms. To ensure the fair comparison, we leverage the StableDiffusion v1.5 checkpoint for all tasks. In parallel with the quantitative results in the main body, DreamSampler achieves better fidelity and data consistency than baselines, as shown in Figure~\ref{fig:inpaint_glass_comp} and \ref{fig:inpaint_smile_comp}.

\subsection{More results for Text-guided Inpainting}

\begin{figure}[h!]
    \centering
    \includegraphics[width=0.93\linewidth]{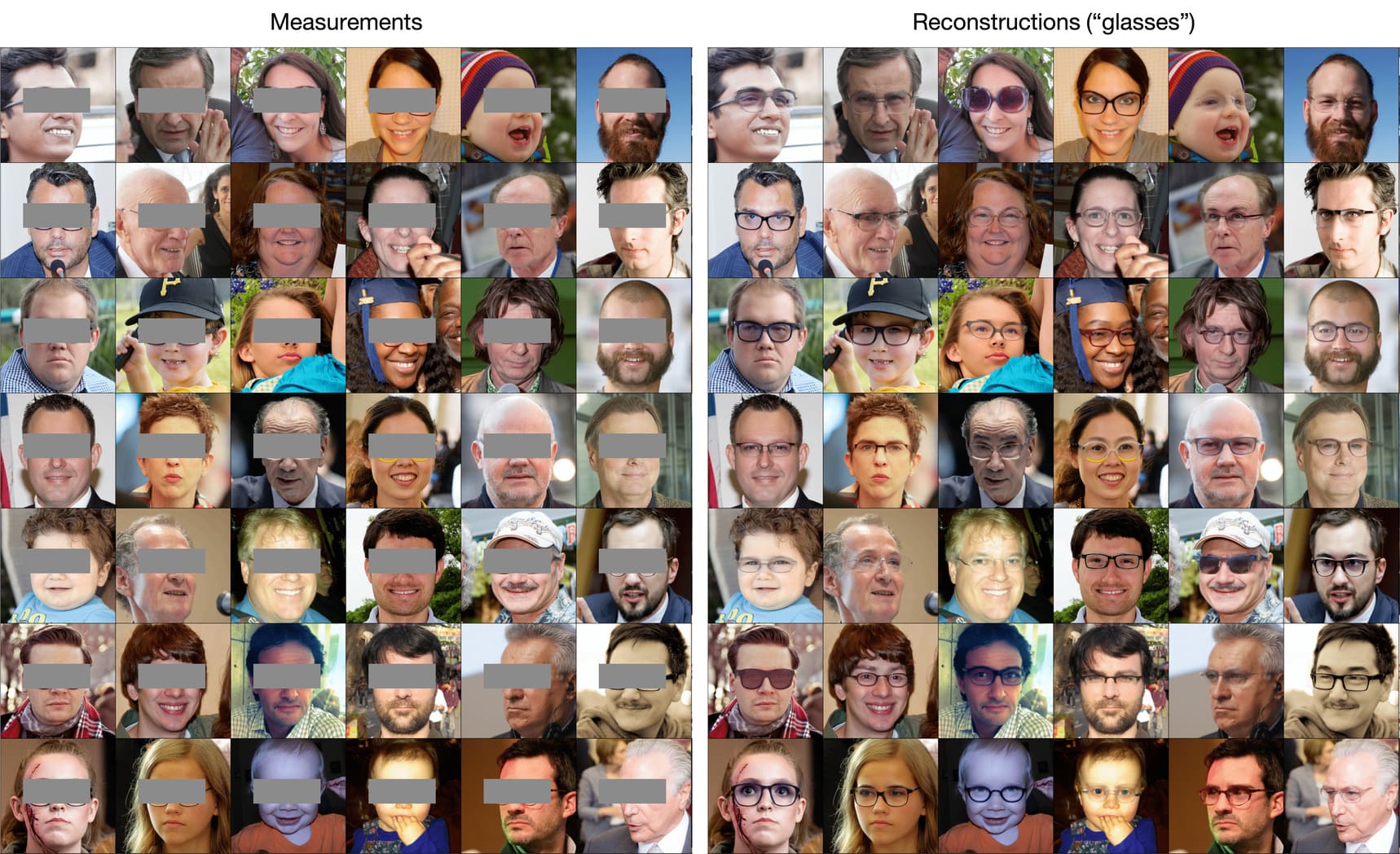}
    \caption{Results for text-guided inpainting with 512x512 FFHQ dataset. Text prompt "wearing glasses" is given.}
    \label{fig:inpaint_glasses}
    \vspace{0.3cm}
    \includegraphics[width=0.93\linewidth]{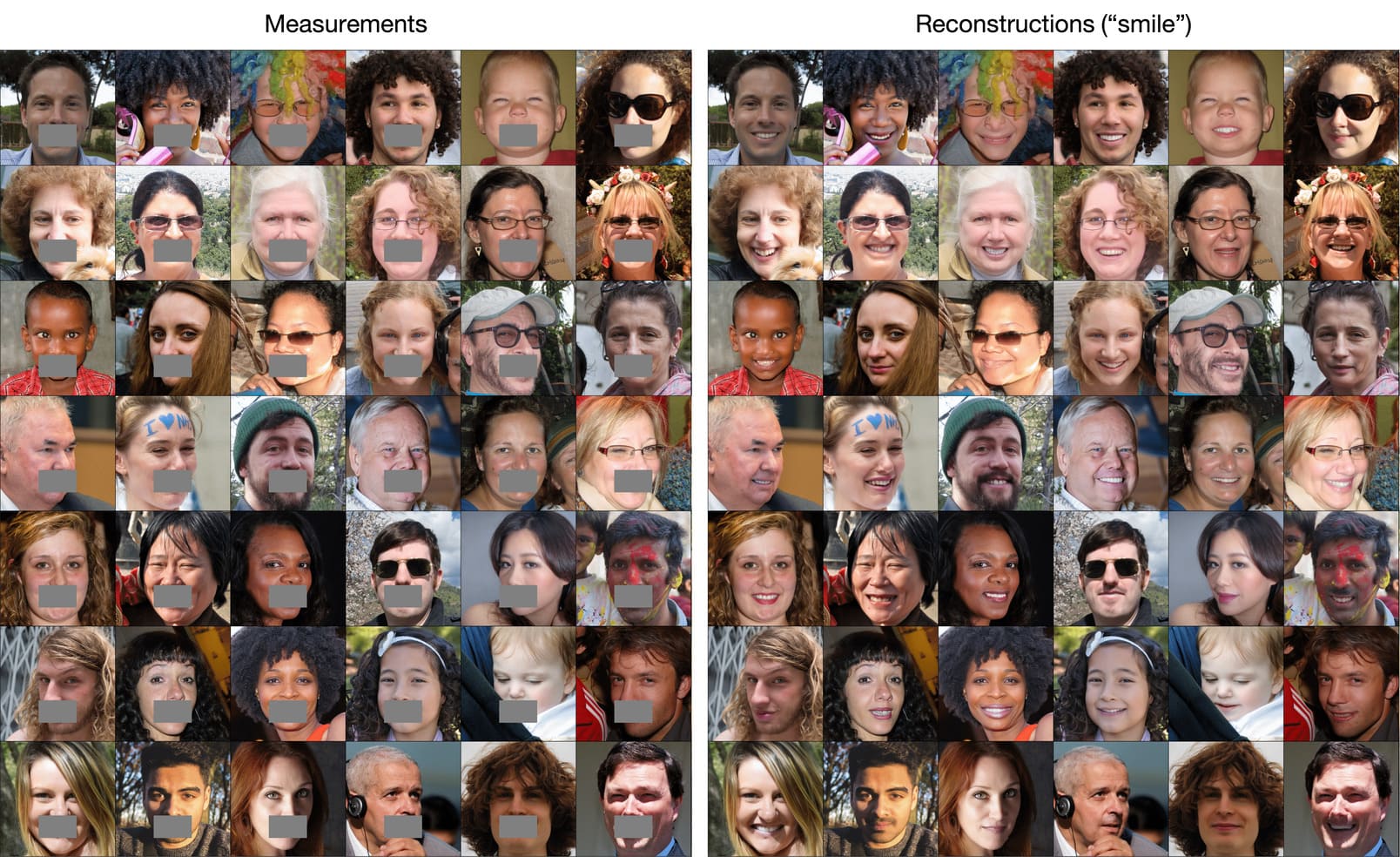}
    \caption{Results for text-guided inpainting with 512x512 FFHQ dataset. Text prompt "smile" is given.}
    \label{fig:inpaint_smile}
\end{figure}

\clearpage
\section{Ablation study}
\subsection{Effect of $\gamma$ in Real Image Editing}
\label{sec:gamma_edit}

\begin{figure}[t]
    \centering
    \includegraphics[width=\linewidth]{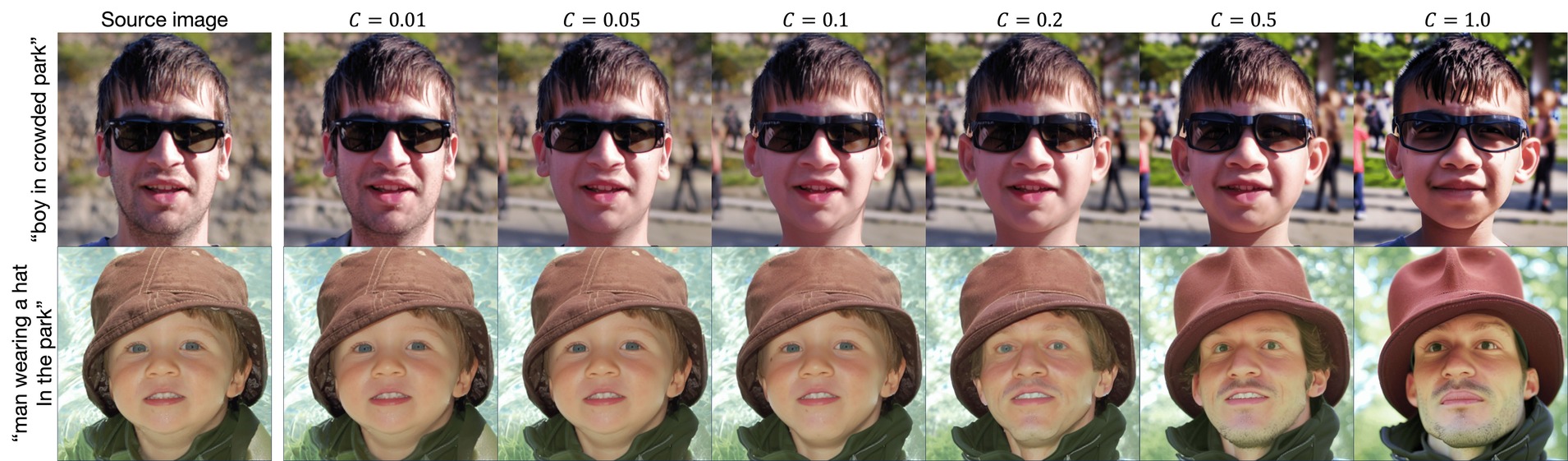}
    \caption{Ablation study on effects of $\gamma$ on real image editing task.}
    \label{fig:abl_gamma}
\end{figure}

From the optimization problem \eqref{eqn:dds_obj}, $\gamma$ could be interpreted as the interpolation weight between $\z_{0|t}(c_\phi)$ and $\z_{0|t}(c_{tgt})$, where $c_\phi$ is utilized for DDIM inversion to initialize $\z_T$.
Thus, setting $\gamma$ close to 1 steers the solution of the problem towards $\z_{0|t}(c_{tgt})$, while setting $\gamma$ close to 0 directs the solution towards $\z_{0|t}(c_\phi)$. Considering the role of $\gamma$ as a scale for the CFG, this interpretation aligns well.
In this study, we use time-dependent $\gamma_t=C\bar\alpha_t$ where $C$ denotes a constant.
To demonstrate the effect of $\gamma$ in a real image editing task, we generated multiple images by varying the value of $C$ from 0.01 to 1.0.
The results in Figure~\ref{fig:abl_gamma} demonstrate that when $C$ is smaller, the generated image closely resembles the reconstruction, while when $C$ is larger, the generated image adheres closely to the text guidance.

\subsection{Effect of Distillation in Image Restoration through Vectorization}
\begin{figure}[t]
    \centering
    \includegraphics[width=\linewidth]{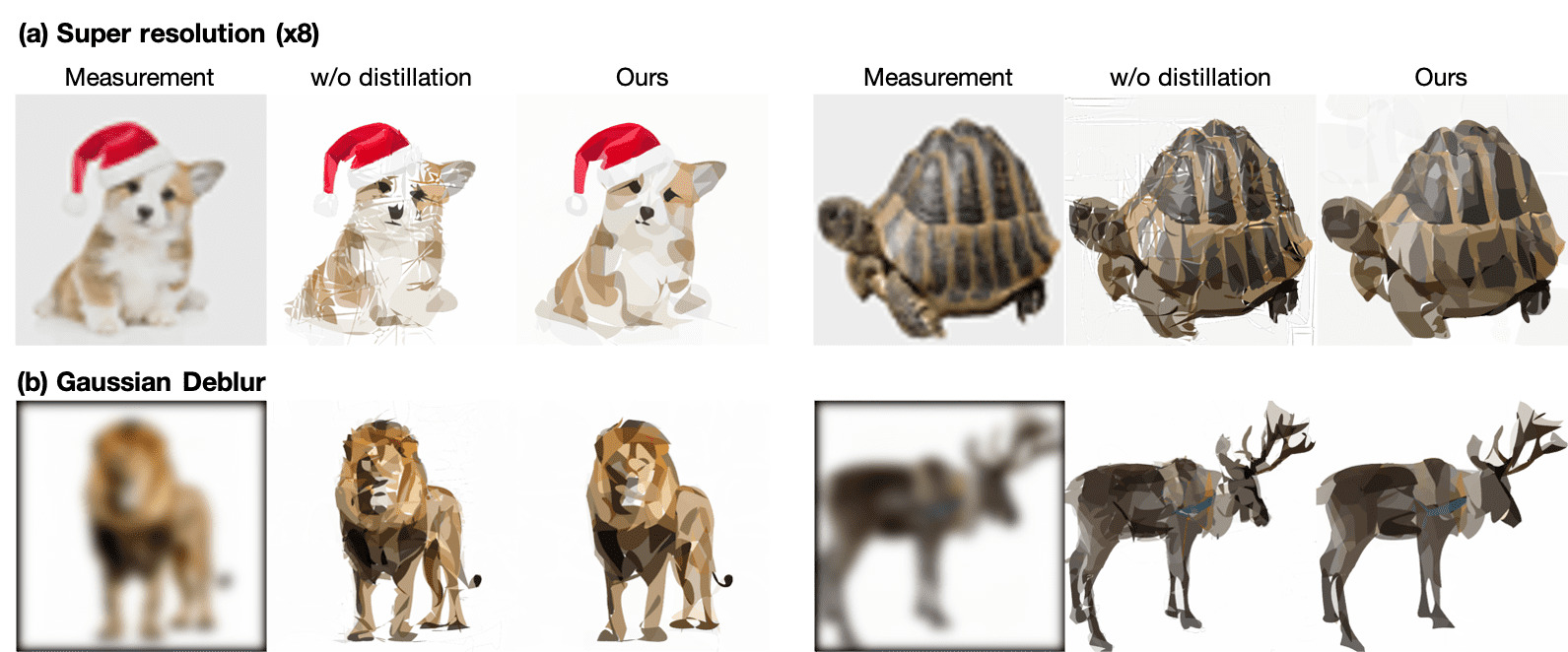}
    \caption{Ablation study on effects of text-guided distillation on restoration task.}
    \label{fig:abl_dis}
\end{figure}

To examine the significance of text-guided distillation in the context of restoration, we adjusted the parameter $\lambda_{SDS}=0$ in Equation (21) of the main manuscript. Figure~\ref{fig:abl_dis} highlights that relying solely on data consistency regularization falls short of achieving a sufficient level of restoration, manifested through the emergence of scattering artifacts within the SVGs. This suggests that the absence of a text-guided distillation refinement process may contribute to severe degradation.

\section{Additional Results}

\subsection{3D representation learning using degraded view}
Since the proposed framework is defined in terms of an arbitrary generator $g$ and generic parameter $\psi$, we apply Dreamsampler to the 3D NeRF \cite{mildenhall2021nerf} representation learning, specifically targeting scenarios with degraded views.
We acknowledge that low-quality, noisy measurements can compromise the detail of 3D modeling, aligning with the motivation of vectorized image restoration task. In this context, the parameters $\psi$ of the generator consist of NeRF MLP, which parameterizes volumetric density and albedo (color). Our goal is to accurately reconstruct NeRF parameters that, when rendered, closely align with the provided degraded view $\y$ and text conditions $c_\y$. 

For this NeRF inverse problem, we consider Gaussian blur operator $\Ac$ with $5 \times 5$ kernel and standard deviation of $10$ following \cite{liu20232}. For a single input image, we introduce novel 16 degraded view images with SyncDreamer \cite{liu2023syncdreamer} and blurring operator $\Ac$.
Then, we first pre-train NeRF with the data consistency regularization $\ell(\y, \Ac g(\psi))$ for warm-up, where $\ell: \mathbb{R}^d \times \mathbb{R}^d \rightarrow \mathbb{R}$ represents a generic loss function.
To facilitate effective representation learning, we integrate $\ell_2$-loss and perceptual similarity loss such as LPIPS \cite{zhang2018unreasonable} for data consistency. Then the latent optimization framework of the NeRF inverse problem is defined as follows:
\begin{equation}
    \min_\psi (1 - \gamma)\lambda_{SDS} \| \Ec_\phi (g(\psi)) - \hat\z_{0|t}(c_{\y}) \|^2 + \gamma \lambda_{DC} \|\y-\Ac g(\psi)\|^2,
    \label{eqn:nerf_optim}
\end{equation}
which is analogous to the SVG inverse problem. We set $\gamma = \bar \alpha_t$ as SVG experiments. Following \cite{zhu2023hifa}, we additionally adapt pixel-space distillation loss to enhance supervision for high-resolution images such as
\begin{equation}
    \| g(\psi) - \Dc_\varphi (\hat\z_{0|t}(c_{\y})) \|^2,
\end{equation}
where $\Dc_\varphi (\hat\z_{0|t}(c_{\y}))$ represents a recovered image via the decoder $\Dc_\varphi$. We further employ $z$-coordinates regularization and kernel smoothing techniques to improve sampling in high-density areas and address texture flickering issues.

\begin{figure}[h!]
    \centering
    \includegraphics[width=\linewidth]{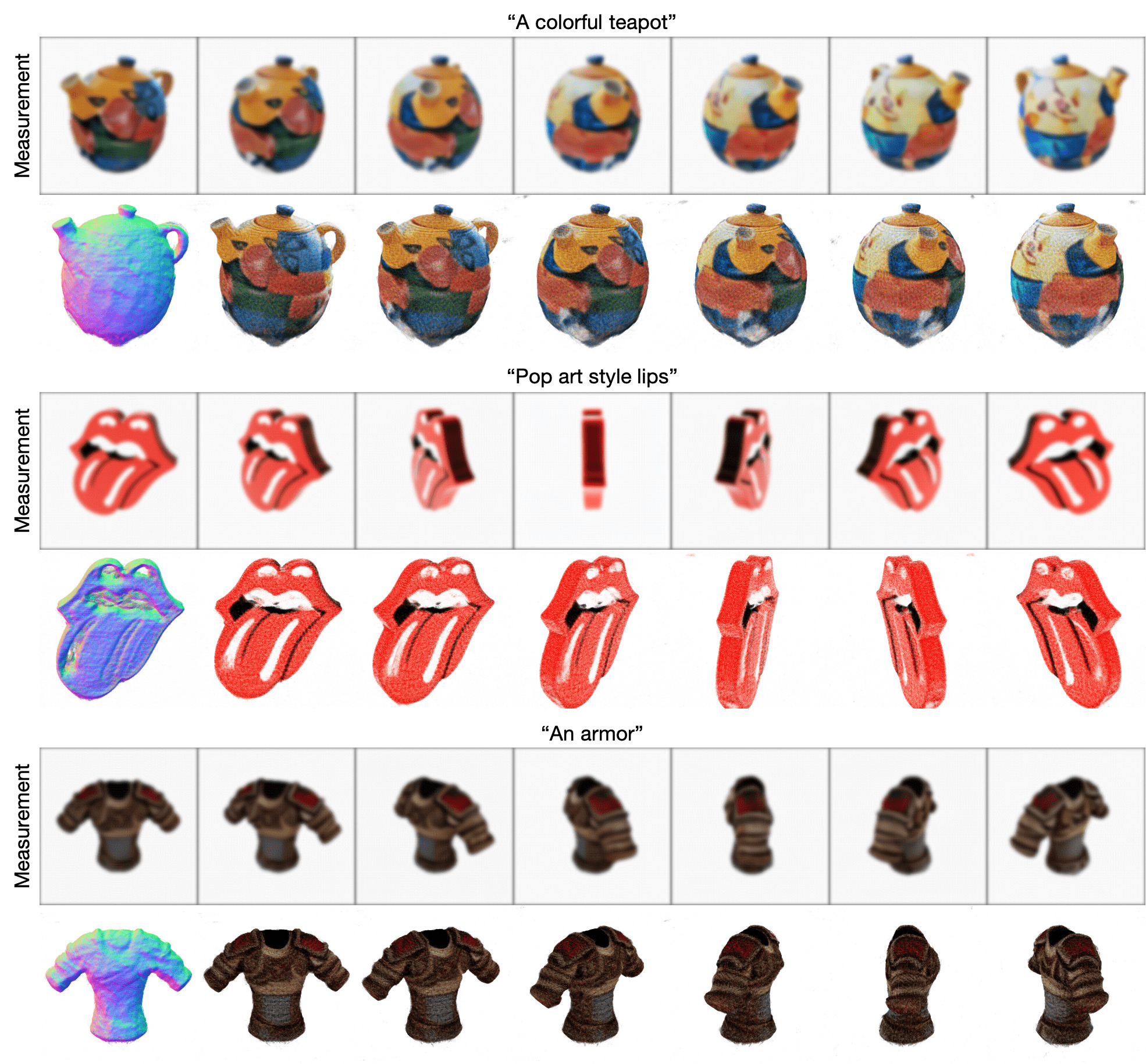}
    \caption{Novel view synthesis via DreamSampler. Blurry views are given.}
    \label{fig:nerf_deblur}
\end{figure}

The findings, as illustrated in \ref{fig:nerf_deblur}, reveal that Dreamsampler successfully achieves high-quality 3D representation despite the compromised quality of the degraded views. Thus, Dreamsampler demonstrates its efficacy in learning 3D representations by directly leveraging degraded views and text conditions, effectively facilitating both distillation and reconstruction processes.

\clearpage
\subsection{More results for Real Image Editing}
\vspace{-0.4cm}
\begin{figure}
    \centering
    \includegraphics[width=0.88\linewidth]{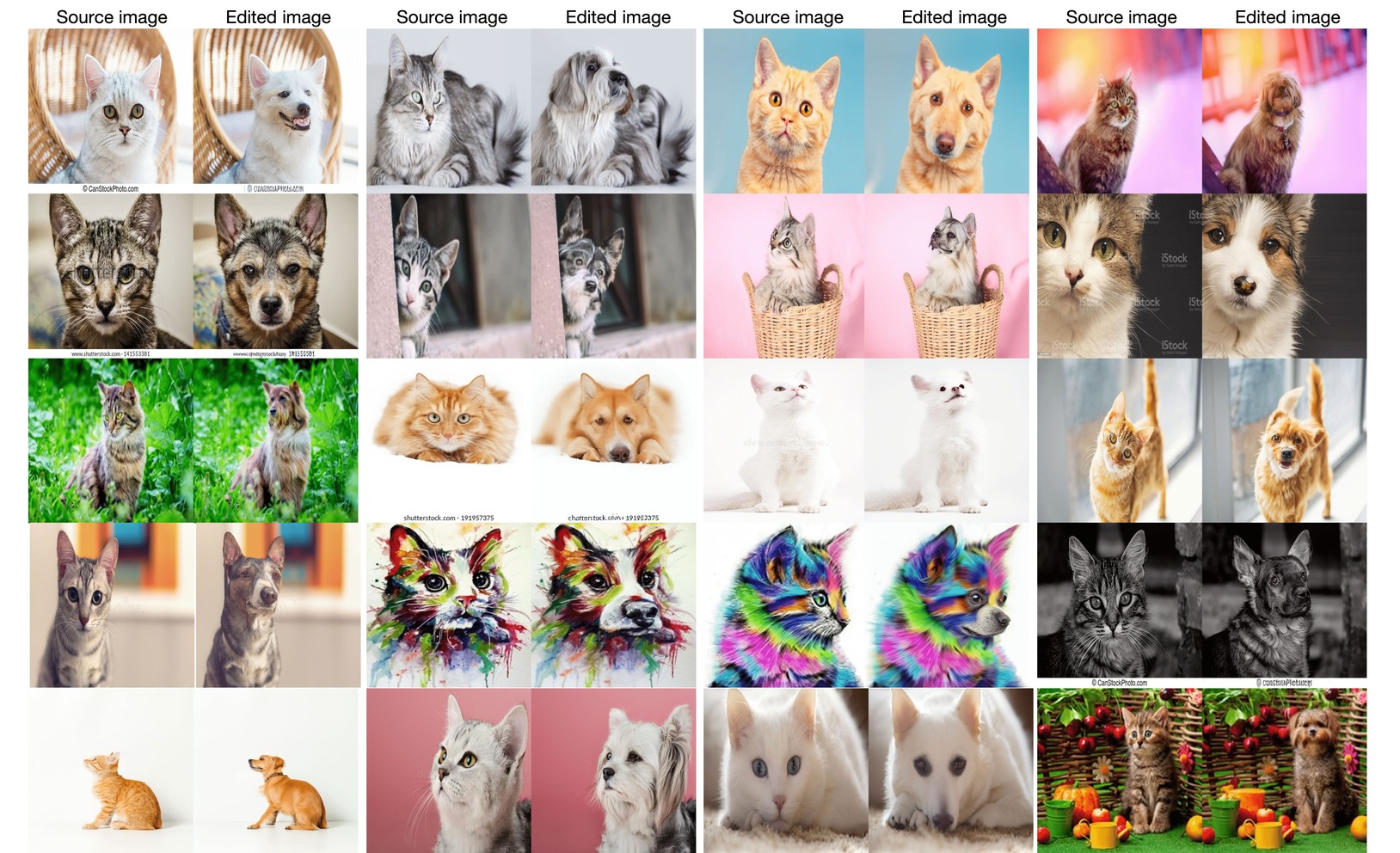}
    \caption{Results for real image editing. Cat $\rightarrow$ Dog. Best views are displayed.}
    \label{fig:edit_c2d}
    \includegraphics[width=0.88\linewidth]{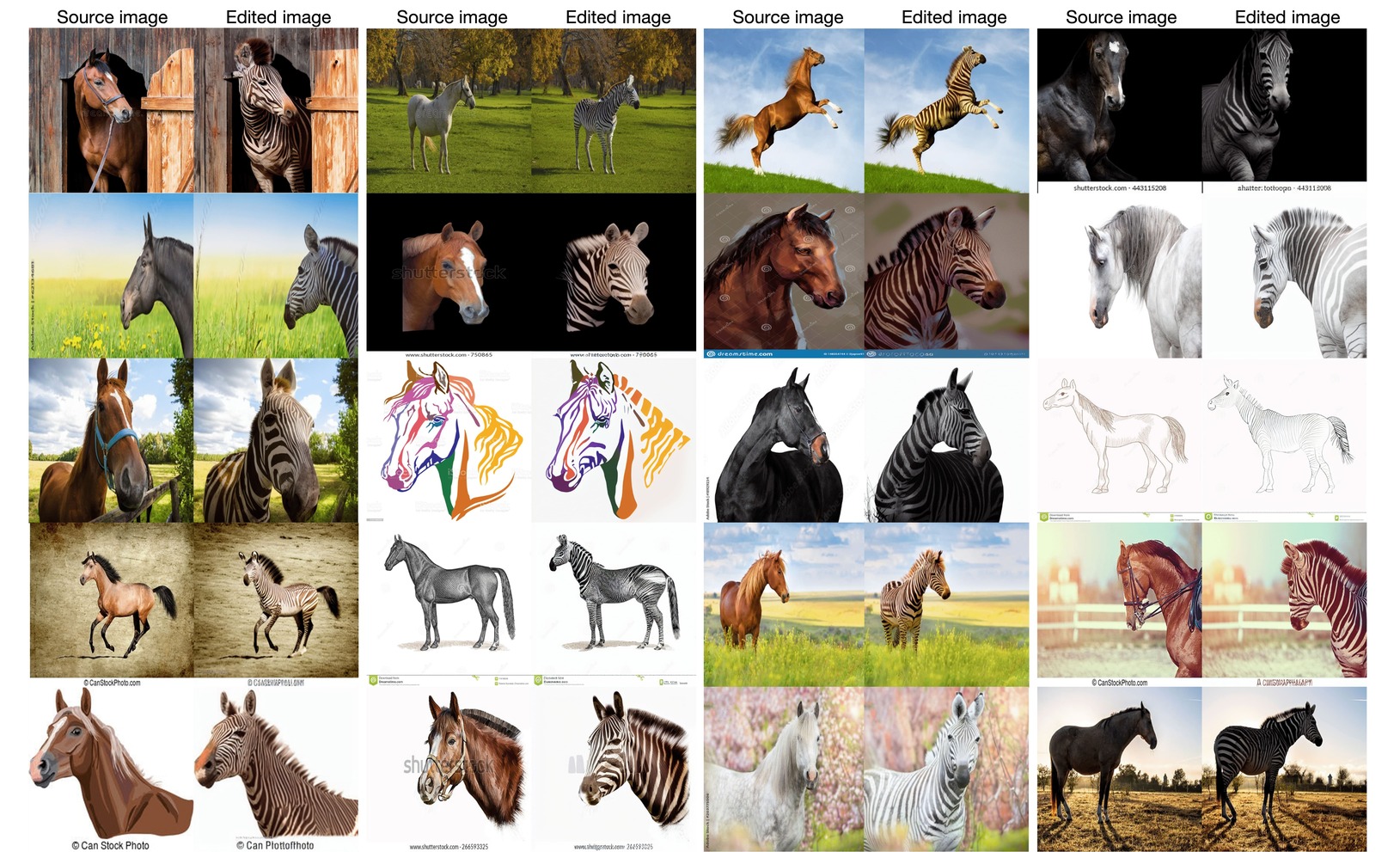}
    \caption{Results for real image editing. Horse $\rightarrow$ Zebra. Best views are displayed.}
    \vspace{-0.9cm}
    \label{fig:edit_h2z}
\end{figure}

\begin{figure}[h!]
\centering
    \includegraphics[width=0.88\linewidth]{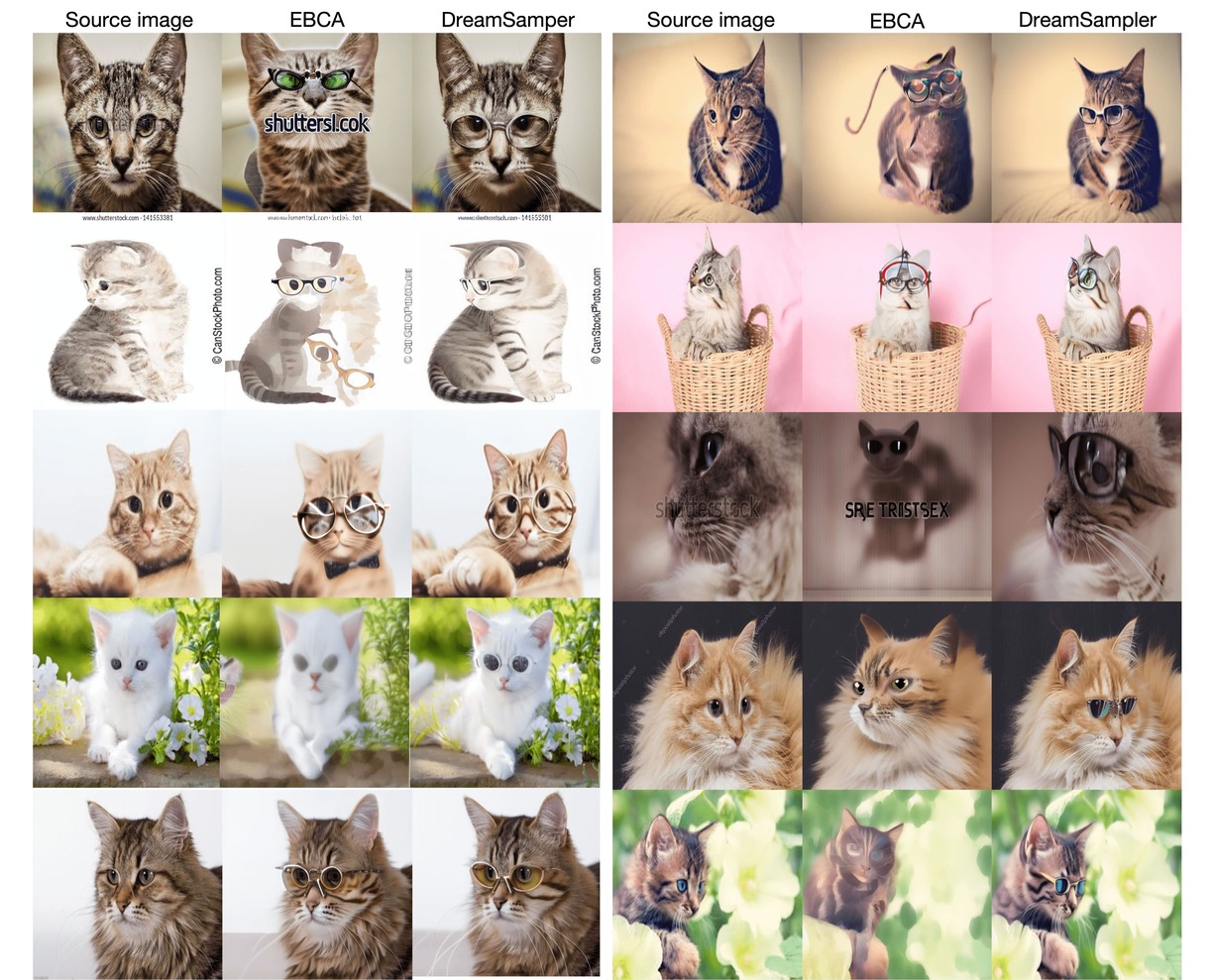}
    \caption{Results for real image editing. Cat $\rightarrow$ Cat with Glasses. Best views are displayed.}
    \label{fig:edit_c2cg}
\end{figure}

\subsection{Text to 3D representation learning with DreamSampler}
\begin{figure}
    \centering
    \includegraphics[width=\linewidth]{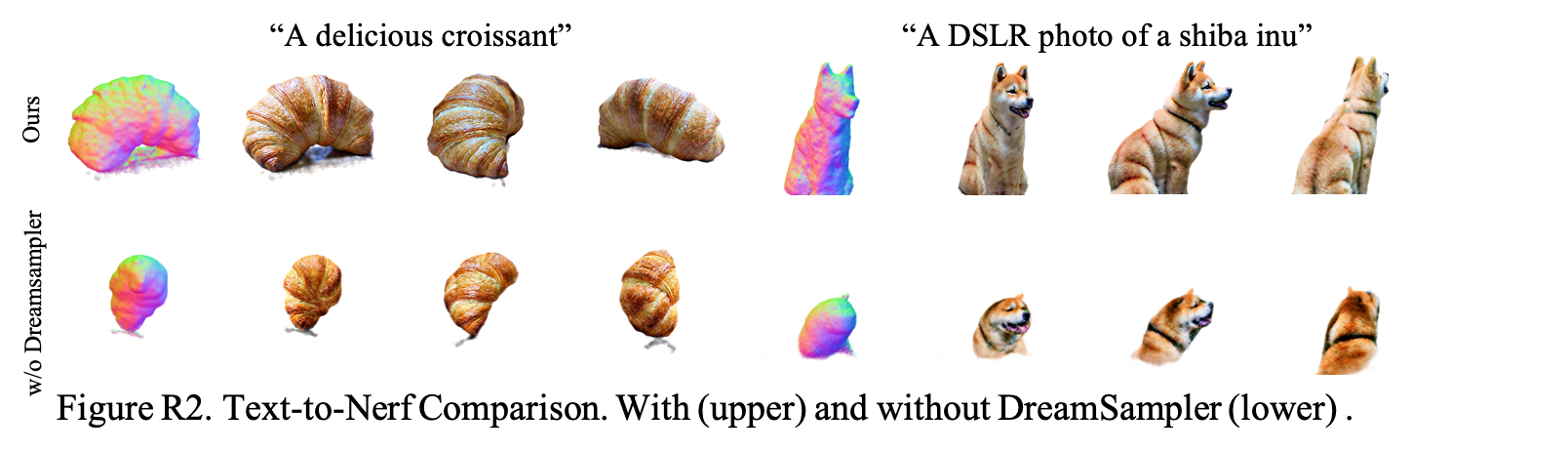}
    \caption{Text-to-NeRF Comparison. With(upper) and without DreamSampler(lower).}
    \label{fig:t23d}
\end{figure}

We demonstrate the generation ability of Dreamsampler with Text-to-3D generation task. We set $\Rc(g(\psi)) = \| g(\psi) - \Dc(\z_{0|t}(c)) \|^2$ as our regularization to improve pixel-level details in high-resolution images as similar to \cite{zhu2023hifa}. We use Adam optimizer with $lr=10^{-3}$ for NeRF weights, optimized for $10^4$ iterations.
Figiure \ref{fig:t23d} shows that Dreamsampler significantly improves the Text-to-NeRF performance with key differences such as decreasing time-step schedule and pixel-domain regularizer $\Rc(g(\psi))$.

\end{document}